\pdfoutput=1
\documentclass[11pt]{article}

\usepackage[preprint]{acl}

\usepackage{times}
\usepackage{latexsym}
\usepackage[T1]{fontenc}
\usepackage[utf8]{inputenc}
\usepackage{microtype}
\usepackage{inconsolata}
\usepackage{graphicx}
\usepackage[dvipsnames]{xcolor}
\usepackage{caption}
\usepackage{amsmath}
\usepackage{amssymb}
\usepackage{booktabs}
\usepackage{placeins}
\usepackage{multirow} 
\usepackage{hyperref}
\usepackage{makecell}  
\usepackage{listings}
\usepackage{fontawesome}

\lstdefinestyle{promptListing}{
    basicstyle=\ttfamily\footnotesize,
    breakatwhitespace=false,         
    breaklines=true,                 
    captionpos=b,                    
    keepspaces=true,                 
    numbersep=5pt,                  
    showspaces=false,                
    showstringspaces=false,
    showtabs=false
}
\title{LLMs Don't Know Their Own Decision Boundaries:\\
 The Unreliability of Self-Generated Counterfactual Explanations} 

\author{
 \textbf{Harry Mayne\textsuperscript{1}},
 \textbf{Ryan Othniel Kearns\textsuperscript{1}},
 \textbf{Yushi Yang\textsuperscript{1}},
 \textbf{Andrew M. Bean\textsuperscript{1}},
\\
 \textbf{Eoin Delaney\textsuperscript{2}},
 \textbf{Chris Russell\textsuperscript{1}},
 \textbf{Adam Mahdi\textsuperscript{1}}
\\
\\
 \textsuperscript{1}University of Oxford,
 \textsuperscript{2}Trinity College Dublin
\\
}

\begin{document}
\maketitle

\begin{abstract}
To collaborate effectively with humans, language models must be able to explain their decisions in natural language. We study a specific type of self-explanation: \textit{self-generated counterfactual explanations} (SCEs), where a model explains its prediction by modifying the input such that it would have predicted a different outcome. We evaluate whether LLMs can produce SCEs that are \textit{valid}, achieving the intended outcome, and \textit{minimal}, modifying the input no more than necessary. When asked to generate counterfactuals, we find that LLMs typically produce SCEs that are valid, but far from minimal, offering little insight into their decision-making behaviour. Worryingly, when asked to generate minimal counterfactuals, LLMs typically make excessively small edits that fail to change predictions. The observed validity-minimality trade-off is consistent across several LLMs, datasets, and evaluation settings. Our findings suggest that SCEs are, at best, an ineffective explainability tool and, at worst, can provide misleading insights into model behaviour. Proposals to deploy LLMs in high-stakes settings must consider the impact of unreliable self-explanations on downstream decision-making. Our code is available at \href{https://github.com/HarryMayne/SCEs}{github.com/HarryMayne/SCEs}.
\end{abstract}

%%%%%%%%%%%%%%%%%%%%%%%%%%%%%%%%%%%%%%%%%%%%%
\section{Introduction}\label{sec:introduction}

\begin{figure*}[t!]
    \centering 
    \includegraphics[width=0.9\textwidth]{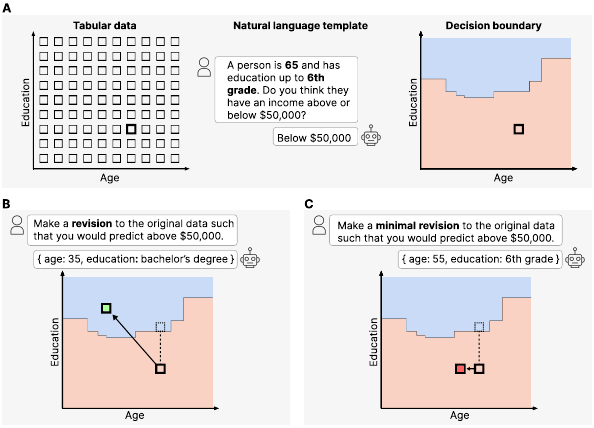}
        \caption{\textbf{Study design.} A. We evaluate models in tabular data, binary classification tasks. The model predictions form a decision boundary across the input space. 
        B. We ask models to provide self-generated counterfactual explanations (SCEs) for their predictions. SCEs are \textit{valid} when they cross the decision boundary (here, red $\rightarrow$ blue) and are \textit{minimal} if they are close to the dashed instance at the decision boundary. When asked to provide counterfactual explanations, we find that SCEs are typically valid but far from minimal.
        C. In separate continuations from the original predictions, we ask models to provide minimal counterfactual explanations. 
        In the majority of cases, these SCEs fail to cross the decision boundary. There is a trade-off between validity and minimality.\label{fig:splash}}
\end{figure*}

Whether LLMs can reliably explain their decisions in natural language has recently become an area of intense research focus~\cite{turpin2023languagemodelsdontsay, parcalabescu-frank-2024-measuring}. Self-explanation is increasingly seen as essential for effective human-computer interaction, allowing users to interrogate model decisions, compare reasoning steps against their prior beliefs, and determine whether behaviours are in line with their goals~\cite{chen2025reasoningmodelsdontsay, baker2025monitoring}. 
One form of self-explanations are \textit{self-generated counterfactual explanations} (SCEs), 
where a model explains its decision by modifying the input such that it would have predicted a different outcome \cite{madsen2024selfexplanations, dehghanighobadi2025llmsexplaincounterfactually}.

Consider the deployment of LLMs to support clinical decision-making~\cite{tu2024conversationaldiagnosticai, arora2025healthbenchevaluatinglargelanguage}. 
A model might predict that a $60$-year-old male with a systolic blood pressure of $135$~mmHg is at high risk of developing heart disease. 
In response, a clinician might ask: \textit{What would need to be different for the model to predict low risk instead?} 
The model could respond with a self-generated counterfactual explanation: 
\textit{If the patient's blood pressure were $110$~mmHg, I would have predicted low risk}. Such explanations highlight the features the model considers important, offer actionable insights for clinicians and patients, and reveal potential flaws in the model. Accordingly, counterfactuals are regarded as a particularly useful form of explanation~\cite{millerExplanationArtificialIntelligence2019, wachter2017counterfactual}.

To serve as effective explainability tools, SCEs must satisfy the following criteria. First, SCEs should be \textit{valid}: the revised input should change the model’s prediction when re-evaluated in a new context window. 
Without validity, the explanation is a misleading representation of the model's counterfactual behaviour. 
Second, SCEs are often expected to be \textit{minimal}: they should make the smallest edit required to change the outcome~\cite{lewis1973counterfactuals, wachter2017counterfactual}. By isolating the changes the model deems consequential, minimal counterfactuals provide clearer insight into the model's decision boundary. In many practical settings, minimality is preferable for satisfying the objectives of explainability~\cite{lipton2018mythos}. For instance, in the heart disease example discussed, a minimal counterfactual would be more actionable for the patient~\cite{keane2020good}.

Prior work has shown that LLMs typically produce valid SCEs but has not addressed minimality~\cite{dehghanighobadi2025llmsexplaincounterfactually, randl2025mind}. The question remains: Can LLMs generate SCEs that are both valid and minimal? To answer this, we use tabular data prediction tasks. In these settings inputs have both natural language and tabular data representations, meaning that we can prompt LLMs in natural language, whilst measuring validity and minimality in structured tabular environments. Through experiments across several LLMs, datasets, and evaluation settings, we show the following:

\textit{Valid but not minimal.} When prompted to generate counterfactual explanations, frontier LLMs achieve near-perfect validity. However, they typically make excessive changes to the original inputs, which trivially flip predictions. These SCEs provide little insight into the models' decision boundaries.

\textit{Minimal but rarely valid.}
When instructed to generate minimal counterfactuals, models generally make overly conservative edits that do not flip their predictions. Such SCEs misrepresent the models' true behaviour and could potentially lead to incorrect downstream decision-making if relied upon in high-stakes deployment. In this regard, LLMs do not know their own decision boundaries. As a consequence, even when using state-of-the-art LLMs, SCEs remain an unreliable, and potentially misleading, method of explaining model behaviour.

%%%%%%%%%%%%%%%%%%%%%%%%%%%%%%%%%%%%%%%%%%%%
\section{Self-generated counterfactual explanations}

Consider a classification task where an LLM $f$ takes natural language input $x$ and predicts a binary output $y$. 
For a given input-output pair, a \textit{counterfactual explanation} is an alternative input $x'\neq x$ that would lead to a specific alternative output $y'\neq y$~\cite{wachter2017counterfactual}. Typically, natural language counterfactuals are identified by algorithms that iteratively perturb keywords in the input (e.g., swapping ``terrible'' $\rightarrow$ ``great'' in a movie review), then test whether each perturbation changes the LLM's decision to the target output~\cite{wang-etal-2024-survey}. In contrast, \textit{self-generated counterfactual explanations} occur when the LLM itself generates a candidate counterfactual to explain its own prediction~\cite{madsen2024selfexplanations, dehghanighobadi2025llmsexplaincounterfactually}. Whether an SCE flips the LLM's prediction can be tested by re-evaluating the LLM in a new context window. Independent evaluation is essential to avoid bias from the presence of the original prediction in context~\cite{dehghanighobadi2025llmsexplaincounterfactually}.

\section{Methods}

In this section, we describe our experiments to evaluate the properties of SCEs.

\subsection{Study design}

Our study design is outlined in Figure \ref{fig:splash}. The core innovation of our approach is to use tabular datasets $\mathcal{Z}=\{z_i\}_{i=1}^N$ containing numerical and ordinal features. We then use natural language templates $\phi$ to convert numerical tabular data $z_i$ to natural language inputs $x_i$ (see §\ref{app:datasets}). As a result, we can prompt the LLMs in the natural language space, but evaluate the SCEs in the tabular data space. This creates a restricted input space, making it tractable to identify the closest point to the original input that would flip the model's prediction, and thus to measure the minimality of the SCEs.

All our datasets have discrete features, and contain all combinations of the feature values. As a result, the datasets are complete in the sense that they cover the full input space. 

For each instance in a dataset, the LLM is first prompted to make a prediction $f(\phi(z_i))=y$; then, given this prediction, it is prompted to revise the input such that it would have predicted the complementary class $y'$. Since the datasets are complete, eliciting predictions over the entire dataset is sufficient to determine the LLM's decision boundary over the input space (Figure \ref{fig:splash}A). 
The resulting decision boundary is then used to assess the properties of the SCE: validity (whether it crosses the decision boundary) and minimality (how much it crosses the boundary by) (Figure \ref{fig:splash}B-C).

\subsection{Datasets}

Our datasets are constructed as simplified versions of real-world datasets. 
In each case, we select two to four discrete features and enumerate all combinations of those features. The datasets are adapted from  \citet{cruz2024evaluating}, \citet{yasserh_housing_prices_2022}, and \citet{heart_disease_45} (for more information see §\ref{app:datasets}). 

\paragraph{Income} Given an individual's age and education level, the LLMs predict whether annual income exceeds $\$50,000$. LLMs are told that the data was collected in 2018 across the US. 
Age is numeric and education is ordinal ($N=1,920$). 

\paragraph{House prices} Given the square‑foot area of a house, the number of bedrooms, bathrooms, and floors, the LLMs predict whether the house price exceeds $\$1,500,000$. LLMs are told that the data was collected in 2015 across the US. All features are numeric ($N=1,600$).

\paragraph{Heart disease} Given an individual's age, sex, systolic blood pressure, and total cholesterol, the LLMs predict whether the individual has coronary heart disease. Age, systolic blood pressure, and total cholesterol are numeric, and sex is binary ($N=1,936$).

\subsection{Models}

\begin{figure*}[t!]
    \centering 
    \includegraphics[width=1\textwidth]{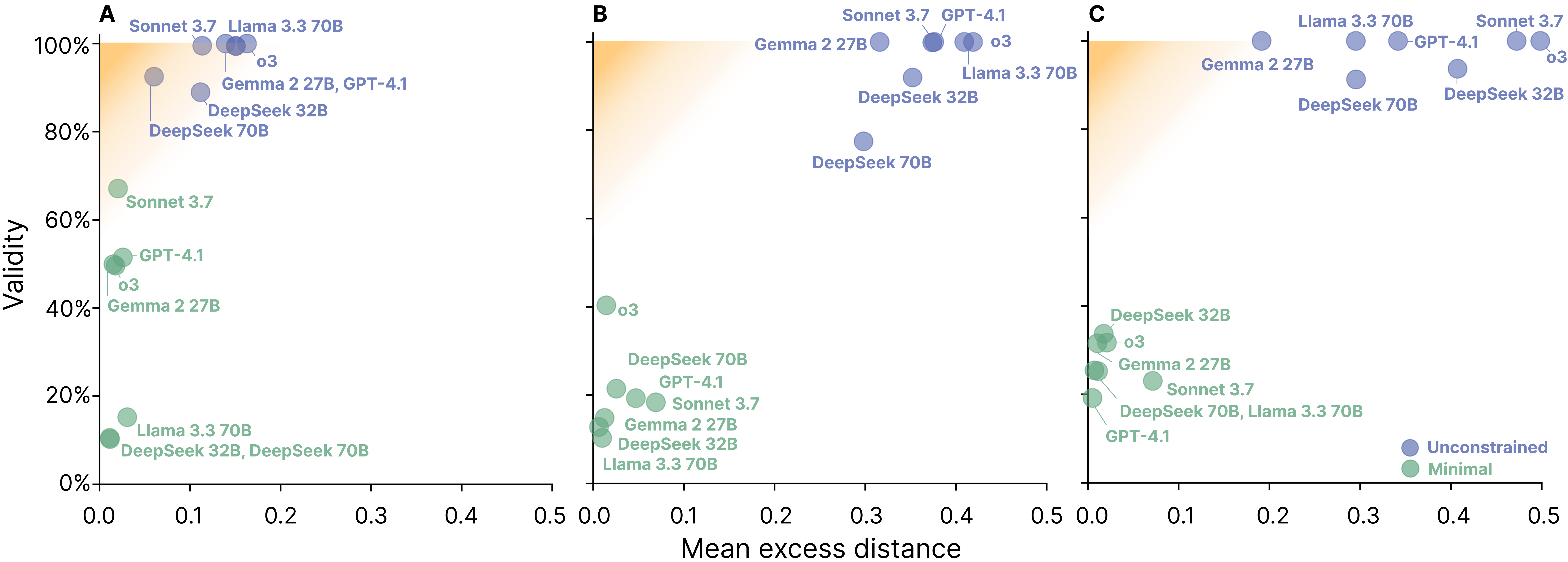}
        \caption{\textbf{SCE validity and minimality for income (A), house prices (B), and heart disease (C)
datasets.} In the unconstrained prompting setting (purple), models typically provide valid SCEs, but they are far from minimal. In the minimal prompting setting (green), validity is notably lower, but, conditional on being valid, minimality is much better. No model can consistently satisfy both criteria across all three datasets. Orange regions indicate the direction of increasing validity and minimality (better performance). \label{fig:scatter_plot}}
\end{figure*}

We evaluate four open-source and three proprietary LLMs: 
Gemma 2 27B \cite{gemmateam2024gemma2improvingopen}, Llama 3.3 70B \cite{grattafiori2024llama3herdmodels}, DeepSeek-R1 Qwen 32B, DeepSeek-R1 Llama 70B \cite{deepseekai2025deepseekr1}, Claude Sonnet 3.7 (non-thinking) \cite{anthropic_claude37_2025}, GPT-4.1 \cite{openai_gpt41_2025}, and o3 \cite{OpenAI2025o3o4mini}. Our main experiments evaluate all models at temperature $0$, except o3, which has a fixed temperature of $1$. 
We discuss the effect of non-zero temperature in §\ref{sec:temperature_1}.

\subsection{Prompt settings}

We evaluate SCEs generated under two prompt settings. First, an \textit{unconstrained} setting, where models are simply asked to produce counterfactual explanations that would flip their predictions. 
This reflects how models behave by default when a user asks them to generate an SCE without any additional constraints. Second, a \textit{minimal} setting, where models are instructed to make the smallest edits necessary to flip their predictions. The prompt wording specifies how minimality is defined and evaluated, ensuring that the models' notion of minimality aligns with our definition (see §\ref{sec:minimality_metrics}). 
To ensure that SCEs remain within the defined input space, we provide models with the feasible ranges of each feature. 
We also enforce a JSON schema with fixed output formats. The full prompts for each setting are provided in §\ref{app:prompting_settings}.

\subsection{Evaluation metrics}\label{sec:metrics}

\paragraph{Validity} For a tabular data instance $z_i$, where $f(\phi(z_i))=y$, the tabular data representation of the SCE $z_i'$ is \textit{valid} if the model flips its decision when re-evaluated in a new context window, i.e. $f(\phi(z_i'))=y'$, where $y'\ne y$. Recall that an SCE is generated for every instance in a dataset. We define \textit{Validity} (Val) as the proportion of dataset instances $z_i\in\mathcal{Z}$ for which the SCE successfully flips the model's prediction.

\paragraph{Minimality}\label{sec:minimality_metrics} 
Any measure of minimality must be defined with respect to an underlying measure of distance.
We use \textit{Gower's Distance} \cite{gower1971general}, a simple pairwise distance function $d(z_i, z_j)$ which is defined over the tabular input space $\mathcal{Z}$ (see §\ref{app:gowers_distance}). This is preferable to more complex functions, e.g. Euclidean distance, as LLMs can reliably calculate it in-context (see §\ref{sec:operationalising_distance}). 
We define the \textit{minimal counterfactual} as the closest point to the original input that would flip the LLM's decision
\[
z_{i,\text{M}} \in \arg\min_{z_i'\in \mathcal{Z}} d(z_i, z'_i), \; \text{s.t.} \; f(\phi(z_i')) = y'.
\]
Given this, we define \textit{Excess Distance (ED)} as the distance from the initial input to the SCE in excess of the distance to the minimal counterfactual, 
\[
\text{ED}_i = d(z_i,z'_i)- d(z_i,z_{i,\text{M}})\,.
\]
We report the mean excess distance over all valid SCEs. This score ranges from $0$ to $1$, where lower scores are more minimal. Since the maximum Gower's Distance across the dataset instances is $1$, this score can be interpreted as a fraction of the span of the dataset.

In addition, we define \textit{Exact Match (EM)} as the proportion of dataset instances $z_i\in\mathcal{Z}$ for which the model exactly identifies a minimal counterfactual.

\begin{table*}[!t]
\centering
\begin{tabular}{lcccccc}
\toprule
\textbf{Model}
  & \multicolumn{3}{c}{\textbf{Unconstrained prompting}} 
  & \multicolumn{3}{c}{\textbf{Minimal prompting}} \\ 
  & Income & House prices & Heart disease 
  & Income & House prices & Heart disease  \\ 
\midrule
Gemma 2 27B  & $0.00$  & $0.00$ & $0.15$ & $31.5$ & $9.94$ & $15.9$ \\
Llama 3.3 70B& $3.33$  & $0.00$ & $0.00$ & $8.80$ & $7.31$  & $10.9$ \\
DeepSeek-R1 32B & $6.89$  & $0.06$ & $0.05$   & $3.94$ & $11.6$  & $19.0$ \\
DeepSeek-R1 70B & $19.7$  & $0.19$ & $0.42$ & $4.22$ & $13.2$ & $17.3$ \\
Claude Sonnet 3.7 & $9.95$  & $0.00$ & $0.00$ & $20.7$ & $9.50$ & $9.92$ \\
GPT-4.1 & $15.7$  & $0.00$ & $0.00$ & $31.1$ & $11.2$ & $10.3$ \\
o3 & $4.27$  & $0.00$ & $0.00$ & $26.2$ & $23.9$ & $17.5$ \\
\bottomrule
\end{tabular}
\caption{\textbf{Exact match: The percentage of SCEs that identify the minimal valid counterfactual across all instances in a dataset.} Models rarely identify the exact minimal counterfactual. Performance is higher in the minimal prompting setting, but no model exceeds 32\%.}
\label{tab:direct_hits}
\end{table*}

%%%%%%%%%%%%%%%%%%%%%%%%%%%%%%%%%%%%%%%%%%%%%%%%%%%%%%%%%%%%%%%%%%%%%%%%%%%%%%%%%%%%%%%%%%%%%%%%%%%%
\section{Results} 
\subsection{Valid but not minimal}

First we consider the unconstrained setting, where models are prompted to generate counterfactual explanations with no additional considerations. Here we find LLMs typically generate valid SCEs, but they are far from minimal (Figure~\ref{fig:scatter_plot}). For example, o3 achieves $100\%$ validity on all three datasets, but has mean excess distance scores of $0.16$, $0.42$, and $0.50$ across the income, house prices and heart disease datasets, respectively. Given an excess distance of $1.0$ is the distance from one extreme of the dataset to the other, a score of $0.16$ can be interpreted as the SCEs overshooting the decision boundary by an average of $16\%$ of the span of the dataset.

Similarly, Table~\ref{tab:direct_hits} shows that models rarely identify the exact minimal counterfactual in this setting, with an average exact match across all models and datasets of only $2.89\%$ (s.d. $5.64$ percentage points).

All LLMs perform notably better in the income dataset. The exact reason for this is unclear, but potentially a result of this dataset being lower-dimensional ($2$ dimensions rather than $4$).

\subsection{Minimal but rarely valid}\label{sec:minimal_prompting_results}

Next we consider the minimal setting, where models are prompted to generate minimal counterfactual explanations. Here we observe a sharp drop in validity (Figure \ref{fig:scatter_plot}). In the income dataset, average validity falls from $97.14\%$ (s.d. $4.49$ pp) to $36.30\%$ (s.d. $23.55$ pp). However, conditional on being valid, the SCEs are more minimal than in the unconstrained prompting setting. For example, whilst o3 now only achieves validity around $40\%$, the mean excess distances of valid SCEs are $0.02$, $0.01$, and $0.02$.

Similarly, Table~\ref{tab:direct_hits} shows that models identify the minimal valid counterfactual more frequently than in the unconstrained setting, with an average exact match across all models and datasets of $14.9\%$ (s.d. $8.01$ pp).

These results suggest a trade-off between validity and minimality, where no model is able to reliably jointly satisfy both criteria. Surprisingly, there is no clear relationship between the general ability of models and their performance on this task, nor does there appear to be any benefit from leveraging inference-time compute to improve reasoning.

\subsection{Where do models place SCEs?}

In the unconstrained setting, we find LLMs often generate SCEs that trivially cross the decision boundary. This behaviour is most obvious in the house price dataset where there are clear monotonic relationships between each feature and average house price (more bedrooms always correspond to higher average price). In this case, models often generate SCEs which maximise or minimise every feature. For example, when modifying the input to predict a house priced above $\$1,500,000$, o3 returns the maximum point in the dataset ($4$ floors, $4$ bathrooms, $5$ bedrooms, area of $10,000$ sqft) in $46\%$ of cases. We find similar behaviours in the heart disease dataset, e.g. o3 selects the maximum or minimum point in $34\%$ of SCEs, but we do not see this pattern in the income dataset. This is potentially because age and education do not necessarily have monotonic relationships with income. However, we do find models favour specific regions. For example, when modifying the input to predict an income above $\$50,000$, Llama 3.3 70B often selects an age between 30 and 50 and education of a Bachelor's degree. 

\begin{figure*}[t]
    \centering 
    \includegraphics[width=0.85\textwidth]{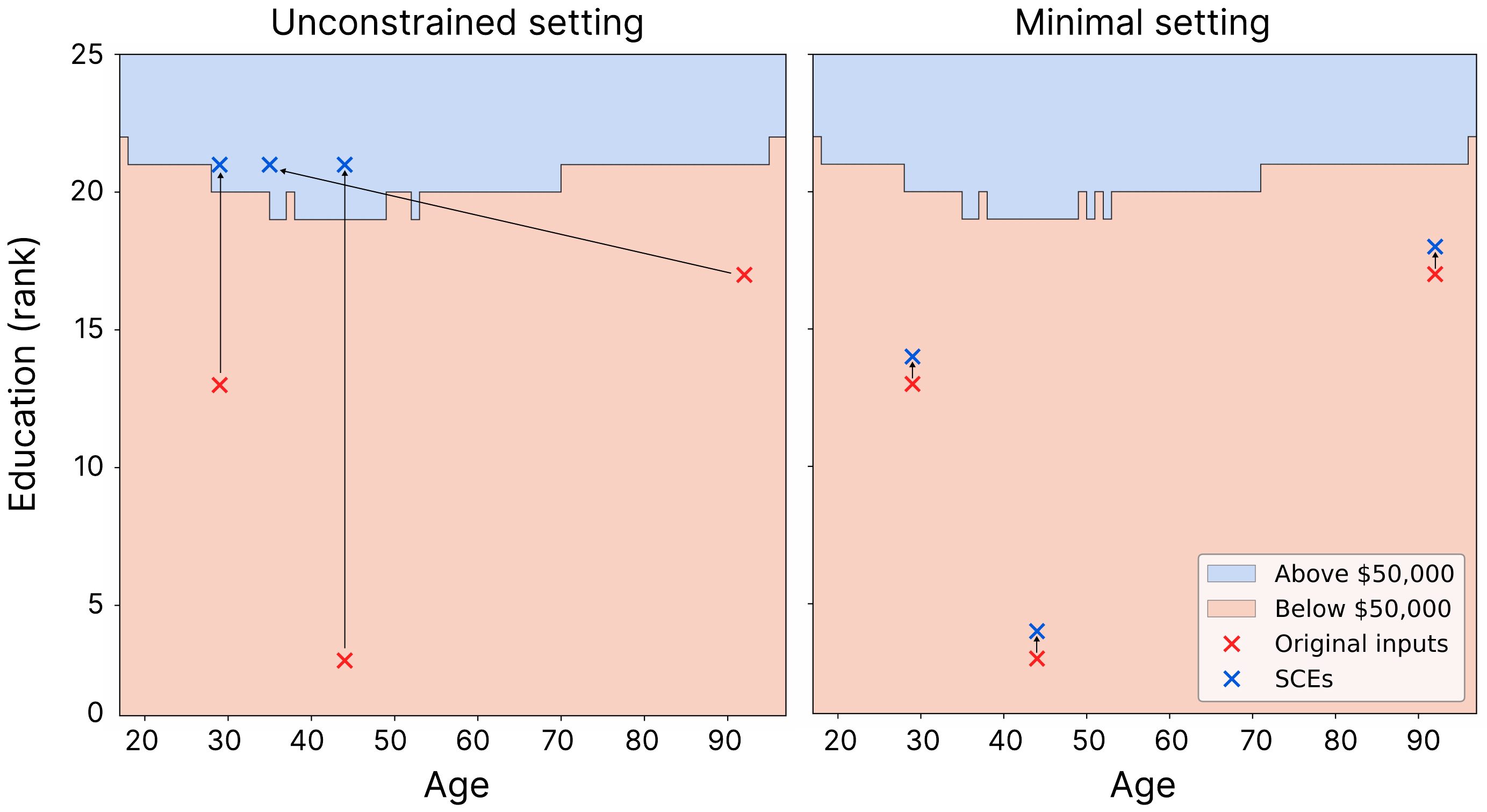}
        \caption{\textbf{Llama 3.3 70B's behaviour on the income dataset.} The three SCEs shown are randomly selected and are representative of the model's general behaviour. In the unconstrained prompting setting, all three SCEs cross the decision boundary and are therefore valid. However, the model makes an excessively large, non-minimal edit to the original input on the far right of the figure. In the minimal prompting setting, the model typically makes overly conservative edits which do not cross the decision boundary. Appendix Figure \ref{fig:density_plots} shows the density of SCEs across all instances in the dataset.\label{fig:main_results}}
\end{figure*}

In the minimal setting, we find LLMs often make overly conservative edits which fail to cross the decision boundary. Figure \ref{fig:main_results} shows representative examples of this for Llama 3.3 70B in the income dataset. Occasionally, this strategy does produce SCEs which cross the decision boundary (if the initial input was close to the boundary), which explains why the SCEs are extremely minimal when valid, but rarely valid.

\subsection{Robustness analysis}

Here we show that the trade-off is robust to changes in the experimental design.

\paragraph{Distance function sensitivity}
We test three alternative distance functions: $L_1$ distance with each feature weighted by its inverse median absolute deviation, $L_2$ distance with each feature weighted by its inverse standard deviation \cite{wachter2017counterfactual}, and semantic distance, calculated as one minus the cosine similarity between inputs encoded using a sentence embedding model \cite{randl2025mind} (see §\ref{app:distance_function_sensitivity} for details). We conduct these experiments with Llama 3.3 70B on the house price dataset since this is the dataset where the difference between the two prompting settings is most pronounced.

Table \ref{tab:robustness-distance} shows the results with excess distance scores normalised by the maximum distance across the dataset so that they fall in the $[0,1]$ range. We find the validity-minimality trade-off occurs across all distance functions.

\begin{table}[h]
\centering
\begin{tabular}{lcccc}
\toprule
 & \multicolumn{2}{c}{\textbf{Unconstrained}} & \multicolumn{2}{c}{\textbf{Minimal}} \\
\cmidrule(lr){2-3}\cmidrule(lr){4-5}
\textbf{Distance} & \textbf{Val}  & \textbf{ED} & \textbf{Val}  & \textbf{ED}\\ %  ($\downarrow$) ($\uparrow$)
\midrule
$L_1$ & $100$ & $0.1403$ & $12.66$ & $0.0169$ \\
$L_2$  & $100$ & $0.1026$ & $11.87$ & $0.0001$ \\
Semantic & $100$ & $0.5845$ & $11.88$ & $0.1280$ \\
Gower & $100$ & $0.1393$ & $15.21$ & $0.0305$ \\
\bottomrule
\end{tabular}
\caption{\textbf{Robustness to distance function choice.} The results are robust across three additional distance functions: $L_1$ distance with features weighted by the inverse median absolute deviation, $L_2$ distance with features weighted by the inverse standard deviation, and a semantic distance defined as one minus the cosine similarity between inputs when embedded by the \texttt{all-mpnet-base-v2} sentence embedding model. Mean excess distances are normalised by the maximum distance between two instances in the dataset. SCEs are generated using Llama 3.3 70B on the house price dataset. Val: Validity. ED: Mean excess distance.}
\label{tab:robustness-distance}
\end{table}

\paragraph{Prompt sensitivity}\label{sec:prompt_sensitivity} We test how sensitive our results are to the specific wording of the SCE-eliciting prompt. For both the unconstrained and minimal settings, we use o3 to generate twenty prompt variations, giving the model detailed instructions on how to modify the original wording (see §\ref{app:prompt_sensitivity}). We then evaluate Llama 3.3 70B’s performance on the house price dataset under each variation.

\begin{figure}[t!]
    \centering 
    \includegraphics[width=0.48\textwidth]{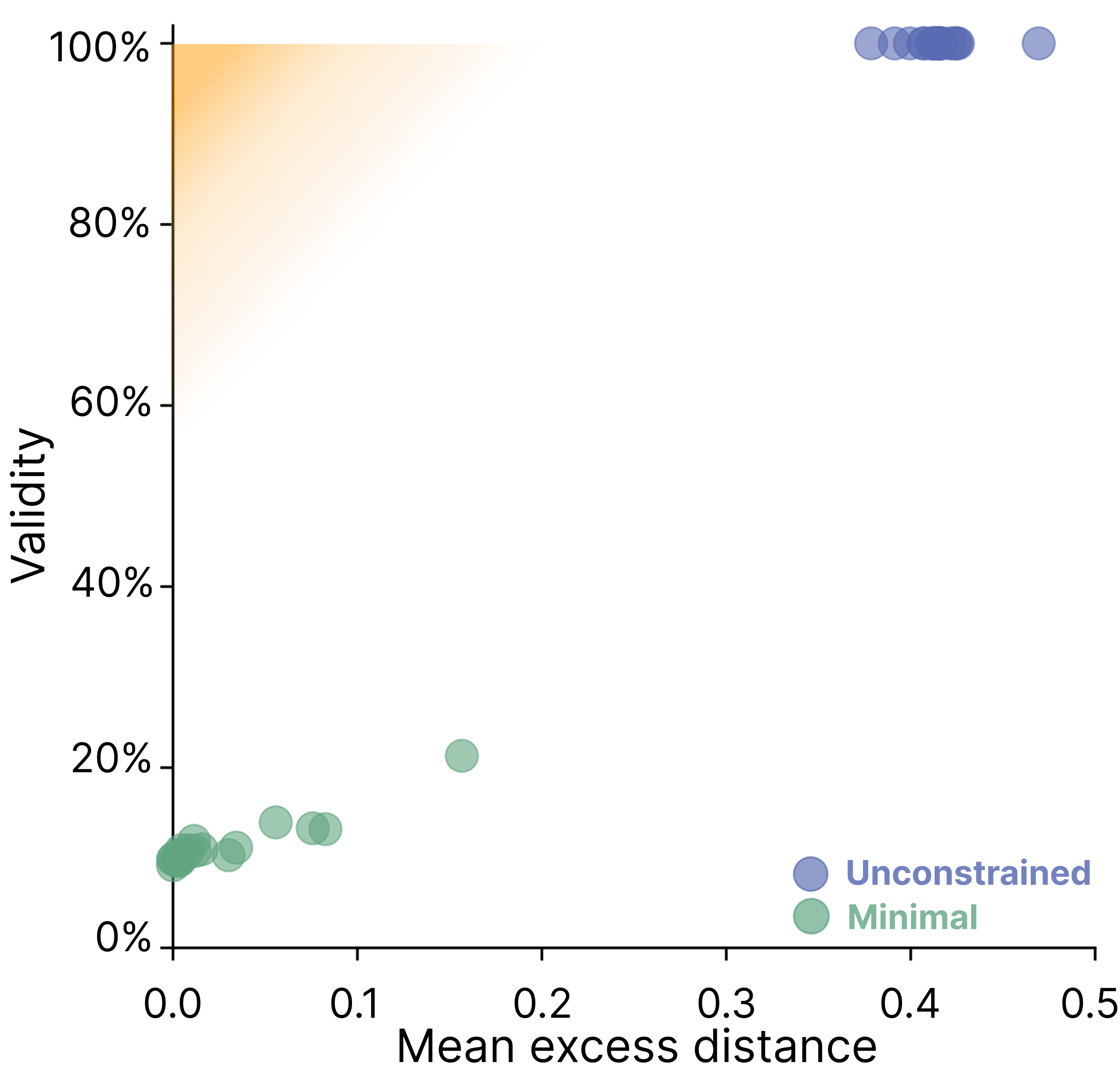}
        \caption{\textbf{Robustness to changes in the SCE-eliciting prompts.} We use o3 to generate 20 versions of both the unconstrained and minimal prompts, then evaluate Llama 3.3 70B's performance on the house price dataset under each variation. Behaviour is robust to the perturbations.\label{fig:prompt_sensitivity}}
\end{figure}

Figure \ref{fig:prompt_sensitivity} shows that performance remains consistent across prompt variations. Behaviour varies only slightly, and importantly, none of the prompts produce SCEs that are both valid and minimal.

\paragraph{Temperature}\label{sec:temperature_1} We test whether our results hold when we use temperature $1.0$ instead of greedy decoding. The results remain consistent: the validity–minimality trade-off persists (see full results in §\ref{app:temp_1}).

\subsection{What limits performance?}

Our results show that no model is able to satisfy both validity and minimality on any dataset. This raises the question: What limits performance? We consider three necessary but not sufficient criteria for success: (i) LLMs must have consistent decision boundaries, (ii) they must be able to operationalise the distance function, and (iii) they must be able to accurately self-predict their behaviour.

\paragraph{Decision boundary consistency}\label{sec:decision_boundary_consistency} Due to the stochasticity of LLMs, the decision boundaries for the prediction tasks may be inconsistent across different generations, making the task of identifying valid and minimal SCEs somewhat ill-defined. To test this, we use o3 to generate $50$ versions of the income task prompt and evaluate the consistency of Llama 3.3 70B's predictions. We elicit predictions at temperature $1.0$ to increase variation (see §\ref{app:prompt_perturbations} for implementation details and example perturbations). We use the income dataset since we can visualise the decision boundary in two dimensions.

\begin{figure}[h]
    \centering 
\includegraphics[width=0.48\textwidth]{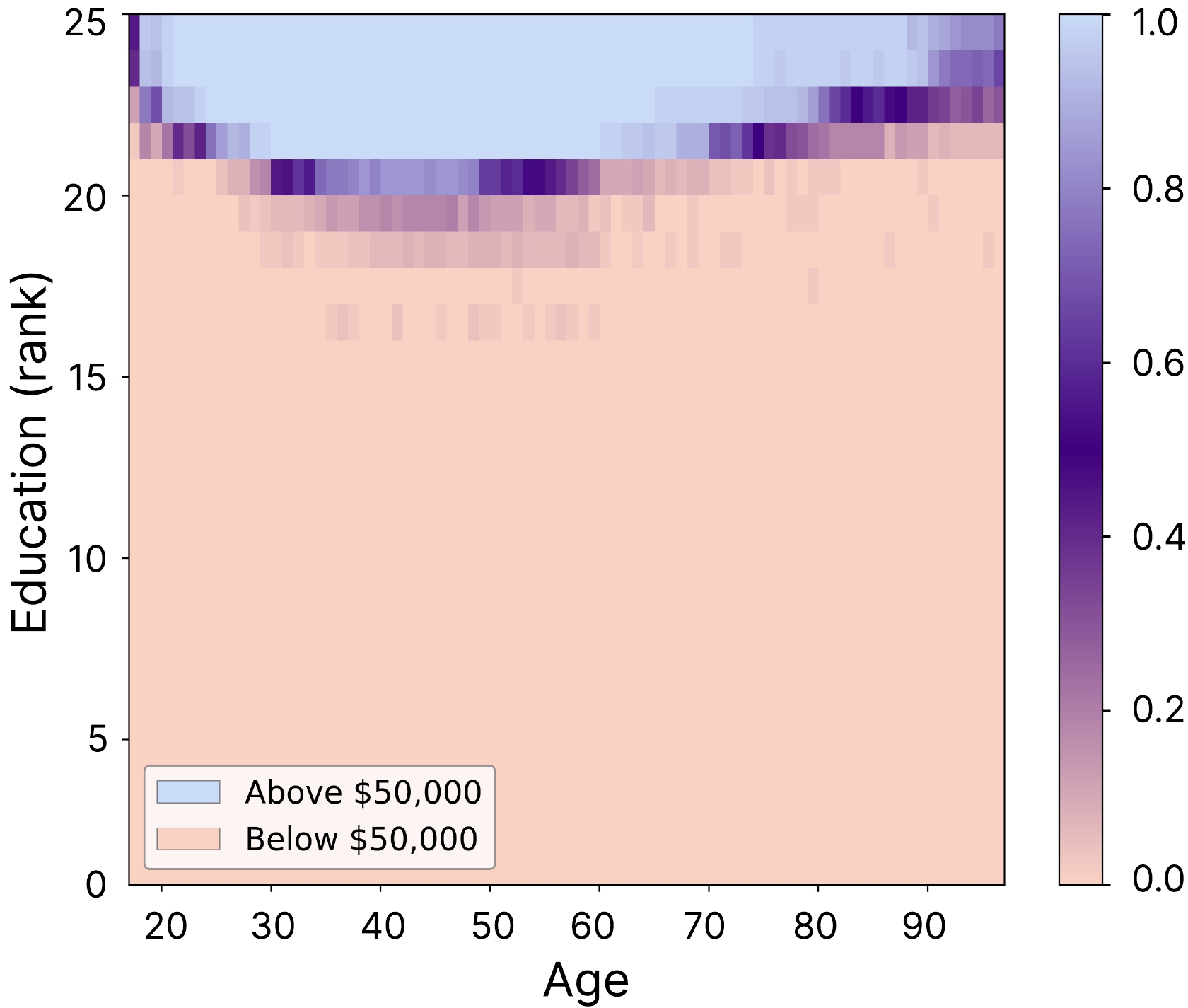}
        \caption{\textbf{Decision boundary consistency across $\mathbf{50}$ prompt perturbations.} Shading indicates the fraction of perturbations where Llama 3.3 70B predicted income above $\$50,000$ for an instance in the dataset. Predictions are elicited at temperature $1.0$.\label{fig:decision_boundary_uncertainty}}
\end{figure}

Figure \ref{fig:decision_boundary_uncertainty} shows that the model's predictions are stable, except in a narrow region around the boundary. Importantly, we find that $91.4\%$ of the original invalid SCEs remain invalid across all $50$ versions of the decision boundary. This suggests that model failures cannot be attributed to decision boundary inconsistency. 

\paragraph{Operationalising distance}\label{sec:operationalising_distance} 
To select the closest valid SCE to the original input, models need to be able to compare candidate SCEs
using the distance function. To do this correctly, they must be able to calculate the pairwise Gower's Distance between each candidate SCE and the original input. To test whether LLMs can do this, we conduct an experiment where models are given an initial instance from the house price dataset and four alternative instances (all randomly selected). Their task is to identify the closest instance to the initial instance using Gower's Distance. We calculate accuracy over $1,000$ unique trials. Random performance is $25\%$ (see §\ref{app_understanding_gower} for details). 

Table \ref{tab:distance_experiment} shows a range in performance. Reasoning models trained to leverage inference-time compute perform notably better ($98-100\%$) than non-reasoning models ($58-72\%$). This reflects general trends in mathematical reasoning benchmarks~\cite{OpenAI2025o3o4mini}. The most capable model we consider, o3, scores $100\%$. 

\begin{table}[h]
\centering
\begin{tabular}{lc}
\toprule
\textbf{Model} & \textbf{Accuracy} (\%) \\
\midrule
Gemma 2 27B    & $58.14$ \\
Llama 3.3 70B  & $71.92$ \\
DeepSeek-R1 32B   & $98.65$ \\
DeepSeek-R1 70B   & $99.79$ \\
Claude Sonnet 3.7   & $70.67$ \\
GPT-4.1   &  $72.25$\\
o3   &  $100.00$\\
\bottomrule
\end{tabular}
\caption{\textbf{Operationalising Gower's Distance.} The rate with which LLMs identify the closest point to an input point among four candidates using Gower's Distance. Accuracy is computed over $1,000$ trials with data coming from the house price dataset.} \label{tab:distance_experiment}
\end{table}

\FloatBarrier

\begin{figure*}[t!]
    \centering 
    \includegraphics[width=1\textwidth]{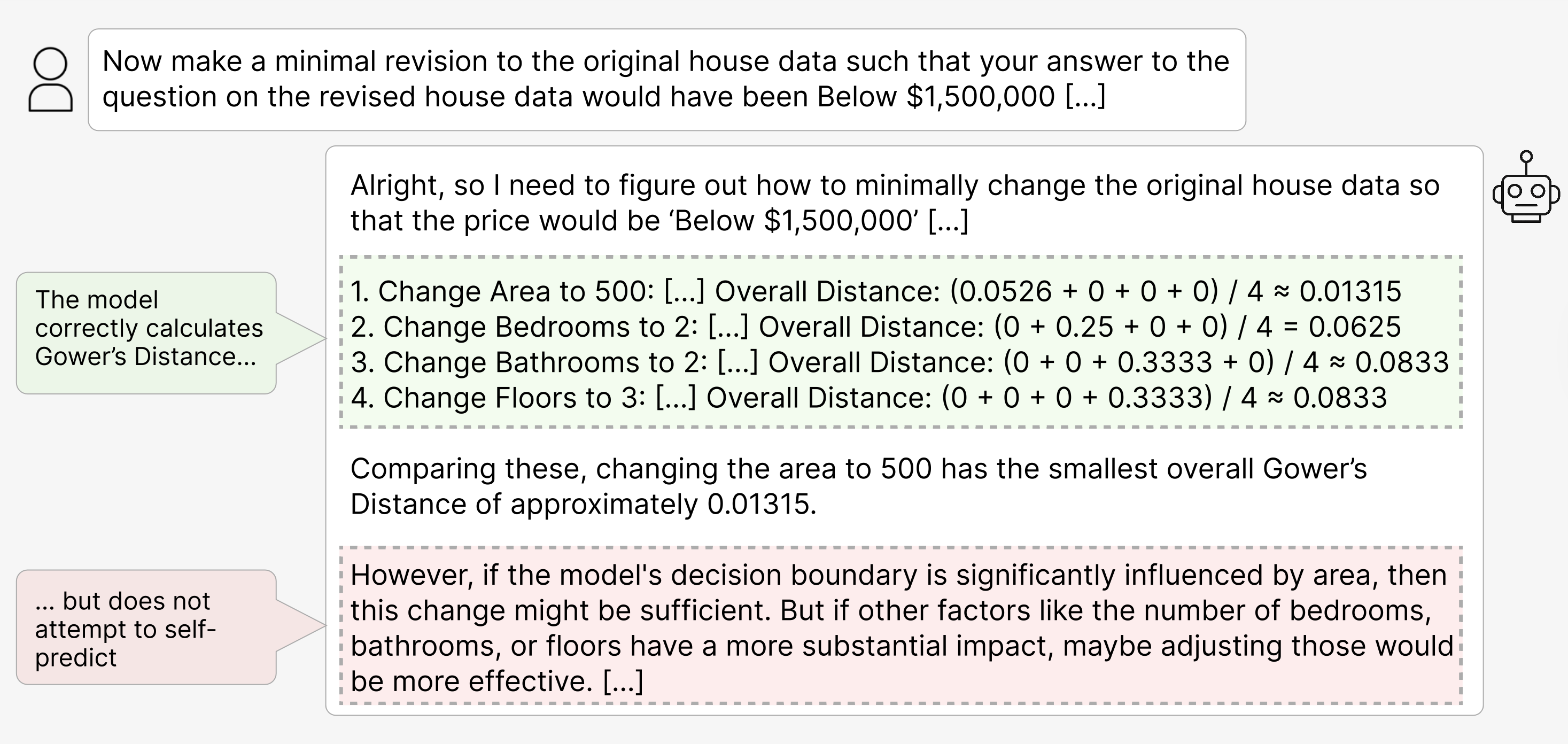}
        \caption{\textbf{DeepSeek-R1 70B can operationalise distance, but does not attempt self-prediction}. DeepSeek-R1 70B leverages Gower's Distance perfectly to compare changes, but does not appear to realise it is predicting its own behaviour. The reasoning traces are taken from an SCE generated on the house price dataset under the minimal prompting setting.\label{fig:reasoning_trace}}
\end{figure*}

Since this experiment is more structured than the real task, we also analyse the distance calculations made by models' when producing the SCEs. By reviewing DeepSeek-R1 70B's reasoning traces, we observe that the model typically approaches the task by first considering multiple candidate SCEs, then calculating the pairwise distance from the initial input to each of the candidates (see an example in Figure \ref{fig:reasoning_trace}). In all cases reviewed, the model correctly calculates Gower's Distance in-context.

Along with the results of the multiple choice experiment, this suggests that the validity-minimality trade-off cannot solely be explained by an inability to operationalise the distance function. While this may be a limiting factor for the weaker, non-reasoning models, it cannot explain why state-of-the-art LLMs fail to produce high-quality SCEs.

\paragraph{Self-prediction}\label{sec:self_modelling} 
Jointly satisfying validity and minimality requires LLMs to accurately predict how they would behave in alternative settings. This could be achieved either through perfect intrinsic knowledge of their decision boundaries or by iteratively testing candidate SCEs and reliably predicting whether each would flip their decision. In either case, we should expect to see evidence that models are attempting to predict their own behaviour, i.e. \textit{self-prediction} \cite{premakumar2024unexpectedbenefitsselfmodelingneural}.

To explore whether models spontaneously engage in self-prediction, we analyse DeepSeek-R1 70B's reasoning traces. While the extent to which chain-of-thoughts faithfully represent intrinsic reasoning remains a subject of debate~\cite{chen2025reasoningmodelsdontsay, barez_chain_2025}, in practice they have proved to contain sufficient information about decision-making to serve as valuable explainability tools~\cite{baker2025monitoring, emmons2025chain}.

By reviewing $30$ randomly-selected reasoning traces from the house price dataset, we find that DeepSeek-R1 70B does not explicitly attempt to self-predict. Although the model shows a good understanding of minimality, it consistently fails to engage with the \textit{self}-explanation aspect of the task. It frequently refers to the decision boundary of an external model, which it is unable to assess, and does not question whether the SCE would flip \textit{its own} decision (see Figure \ref{fig:reasoning_trace}). This is a fundamental necessary condition that is not met.

We also consider an experiment where we explicitly prompt models to self-predict in their reasoning traces. We instruct models to follow a six-step plan encouraging them to consider candidate SCEs, make explicit self-predictions and update their candidate to converge on the minimal valid counterfactual (see §\ref{app:self_prediction_ablation} for details). Evaluating DeepSeek-R1 70B on the house price dataset we qualitatively observe the model making self-predictions in-context and having a better understanding of the self-explanation requirements of the task. Despite this, the aggregate results are largely unchanged with validity increasing from $21.44\%$ to $23.34\%$, at the cost of excess distance increasing from $0.025$ to $0.033$. Anecdotally, we also observe that the self-predictions are often incorrect.

Whether accurate self-prediction is a fundamental limitation of LLMs is unclear and a question for future research. This would perhaps be unsurprising since standard pre-training techniques only incentivise acquisition of knowledge about the external world~\cite{radford2018improving} and common post-training methods, including those using reinforcement learning to leverage inference-time compute~\cite{openai2024learning, deepseekai2025deepseekr1}, never reward models for accurately predicting how they would behave in independent context windows. In short, there is no optimisation pressure to develop a self-model \cite{premakumar2024unexpectedbenefitsselfmodelingneural}. New learning objectives are likely required to incentivise accurate self-prediction.

%%%%%%%%%%%%%%%%%%%%%%%%%%%%%%%%%%%%%%%%%%%%%%%%
\section{Related work}

We discuss two broad related areas: counterfactual explanations and self-explanations.

\paragraph{Counterfactual explanations}
Counterfactuals have been proposed as a way of explaining opaque models without requiring an understanding of their internal mechanisms \cite{wachter2017counterfactual}. In NLP settings, counterfactuals are typically generated using keyword-based perturbation methods~\cite{wang-etal-2024-survey} or using a separate language model to produce candidate counterfactuals \cite{wu2021polyjuice, nguyenLLMsGeneratingEvaluating2024}.

Philosophical work has long suggested that counterfactual explanations should capture the \textit{closest possible world}~\cite{lewis1973counterfactuals}, since this is often more actionable for the user. This sparked the importance of minimality in subsequent algorithmic research \cite{wachter2017counterfactual}, and led to the deployment of optimisation methods that minimise the distance between the original instance and the candidate counterfactual \cite{keane2020good}. In NLP settings, the properties of counterfactuals have been well studied~\cite{wang-etal-2024-survey, nguyenLLMsGeneratingEvaluating2024, mcaleese2024comparativeanalysiscounterfactualexplanation}. However, there is limited work measuring minimality in natural language domains. This is because, while it is easy to define \textit{distance} in natural language spaces, measuring \textit{minimality} requires identifying the minimal counterfactual, which is computationally challenging given the input space is unconstrained. Our work addresses this by using tabular datasets to create restricted input spaces, allowing us to precisely measure minimality.

\paragraph{Self-explanations and SCEs}
As LLM capabilities have progressed, the ability of LLMs to offer explanations of their own decisions has emerged as a potential new paradigm for explainability~\cite{madsen2024interpretabilityneedsnewparadigm}. 
LLMs can provide plausible self-explanations either through their chain-of-thoughts or by explaining themselves post-hoc \cite{pmlr-v235-chen24bl}. Many studies implicitly treat these as interpretability tools~\cite{barez_chain_2025}; however, the critical question is whether these explanations are faithful representations of models' internal reasoning~\cite{wiegreffe-etal-2021-measuring, turpin2023languagemodelsdontsay, lanham2023measuringfaithfulnesschainofthoughtreasoning, parcalabescu-frank-2024-measuring, siegel-etal-2024-probabilities, barez_chain_2025}.

An SCE is a model's attempt to self-explain using a counterfactual explanation. This is a structured form of self-explanation since there are additional constraints placed on the writing process compared to free-text explanations \cite{wiegreffe2021teach}. Self-generated counterfactuals were first introduced by~\citet{madsen2024selfexplanations}, and later evaluated in subsequent works~\cite{dehghanighobadi2025llmsexplaincounterfactually, randl2025mind}. These studies primarily focus on measuring validity across a range of open-ended generation settings. Our work builds on this by evaluating both validity and minimality, and by examining the trade-off between them.

%%%%%%%%%%%%%%%%%%%%%%%%%%%%%%%%%%%%%%%%%%%%%%%%%%%%%%%%%%%%%%%%%%%%%%%%%%%%%%%%%%%%%%%%%%%%%%%%%%%%
\section{Conclusion}

Our findings reveal an important limitation of self-generated counterfactual explanations: a trade-off between validity and minimality. At best, SCEs are an ineffective means of explaining model behaviour, since LLMs naturally provide trivially valid counterfactuals. At worst, SCEs provide misleading insights into model behaviour, which could incorrectly steer human decision-making, potentially causing harm. Proposals to deploy LLMs in high-stakes settings must consider the impact of unreliable self-explanations.

Our study also investigates the factors that might limit performance, finding that models struggle to recognise the \textit{self-}explanation elements of this task. Crucially, they do not spontaneously engage in self-prediction, a necessary condition for strong performance. Furthermore, performance does not significantly improve when models are explicitly prompted to self-predict in their reasoning traces, raising an important question about whether this is a fundamental limitation of LLMs.

\section{Limitations} 

\paragraph{Alternative minimality metrics}
Our primary minimality metric is mean excess distance (§\ref{sec:minimality_metrics}). This metric is designed to capture the minimality of \textit{valid} SCEs, but does not account for the properties of invalid SCEs. In some cases, it might be desirable to understand the distribution of these SCEs too. For example, we might wish to know whether the invalid SCEs lie just the wrong side of the decision boundary or are far removed. By introducing mean excess distance, we hope future work might design new metrics that incorporate both valid and invalid SCEs.

\paragraph{Self-prediction capabilities} 
Our analysis suggests that models do not spontaneously engage in self-prediction when generating SCEs. We also show that aggregate performance is largely unchanged when we explicitly prompt LLMs to self-predict in their reasoning traces. However, we do not conduct detailed analysis into the self-prediction accuracy, make claims about whether LLMs are fundamentally capable of this, or the mechanisms by which they might hypothetically achieve this. These questions relate to the metacognitive capabilities of LLMs~\cite{steyvers2025metacognitionuncertaintycommunicationhumans} and are promising directions for future work. When models are prompted or fine-tuned to self-predict within their chain-of-thoughts, can they do so accurately? If so, does SCE performance improve? Or are models fundamentally unable to accurately self-predict? Appendix \ref{app:self_prediction_ablation} discusses these thoughts further.

\paragraph{Generalisation to real-world datasets}
Our experiments use three synthetic datasets, designed as simplified versions of real-world datasets. This is central to our methodology, as it allows us to generate datasets which contain every combination of the discrete input features. As a result, the exact minimum counterfactual can be located, enabling the measurement of SCE minimality. An additional benefit of using toy datasets is that they provide the simplest possible environment in which models might identify valid and minimal counterfactuals. Across all experiments, we use datasets with two to four input dimensions. Since models fail in these datasets, we anticipate that they will also struggle in more complex, higher-dimensional datasets, where locating a valid and minimal counterfactual is significantly harder.

\section*{Acknowledgements}

We thank the reviewers for their helpful feedback. We also thank Jabez Magomere and Jakob Foerster for insightful conversations and feedback during the research process. Harry Mayne’s PhD is supported by the Economic and Social Research Council grant ES/P000649/1. Andrew Bean’s PhD is supported by the Clarendon Fund Scholarships at the University of Oxford. 
Yushi Yang's PhD is supported by the Oxford Internet Institute and the Dieter Schwarz Foundation Project Award. 
Ryan Othniel Kearns was supported by a Fellowship from the Cosmos Institute.

\bibliography{custom}

\appendix

%%%%%%%%%%%%%%%%%%%%%%%%%%%%%%%%%%%%%%%%%%%%%%%%%%%%%%%%%%%%%%%%%%%%%%%%%%%%%%%%%%%%%%%%%%
\section{Main experiment implementation details}\label{app:methodology}

%%%%%%%%%%%%%%%%%%%%%%%%%%%%%%%%%%%%%%%%%%%%%%%%%%%%%%%%%%%%%%%%%%%%%%%%%%%%%%%%%%%%%%%%%%
\subsection{Datasets}\label{app:datasets}
%%%%%%%%%%%%%%%%%%%%%%%%%%%%%%%%%%%%%%%%%%%%%%%%%%%%%%%%%%%%%%%%%%%%%%%%%%%%%%%%%%%%%%%%%%
For each of the three datasets used, the full list of features, and the discrete values they can take, is given in Table \ref{tab:dataset_features}. Each dataset is synthetically generated using all possible feature combinations. This ensures that the datasets completely represent the input space. 

\paragraph{Income} For the income dataset task, the model is provided with an individual's age and education level and must predict whether their income exceeds $\$50,000$. The model is told that the data is collected from United States residents in $2018$. Ages are restricted to integers between $17$ and $96$, and education takes one of $24$ ordinal values (see Table~\ref{tab:dataset_features}). This dataset is inspired by the Folktexts dataset~\cite{cruz2024evaluating}, though our data is simpler, only using two of the features in this dataset. A total of $80\times 24$ feature values yields $1,920$ distinct instances in the dataset.

\paragraph{House prices} For the house price dataset task, the model must predict whether a house is priced above or below $\$1,500,000$ given the number of bedrooms, bathrooms, floors, and its size in square feet. The models are told that the data was collected in $2015$ from houses across the United States. The number of bedrooms ranges from $1$ to $5$, bathrooms and floors from $1$ to $4$, and square footages, from $500$ to $10,000$ (in steps of $500$). This dataset is inspired by popular house price prediction datasets common in traditional machine learning, e.g. \citet{yasserh_housing_prices_2022}. Our dataset is generated by considering all unique combinations of the discrete features. A total of $5\times 4\times 4\times 20$ feature values yields $1,600$ distinct instances in the dataset.

\paragraph{Heart disease} For the heart disease dataset task, the model must predict whether a patient has heart disease from their age, sex, systolic blood pressure (mmHg), and total cholesterol (mg/dL). Sex is a categorical variable. For age, systolic blood pressure, and total cholesterol, we provide the model with lists of representative ordinal values (details in Table \ref{tab:dataset_features}). This dataset is inspired by the UCI Machine Learning Repository dataset \cite{heart_disease_45}. A total of $11\times 2\times 8\times 11$ feature values yields $1,936$ distinct instances in the dataset.

A question that may appear important is whether models can accurately do these prediction tasks with respect to real world data. For example, can models accurately predict incomes in $2018$, United States. This turns out to be irrelevant for evaluating SCEs. This is because the ground truth for SCE experiments is the models' initial predictions themselves. As a result, issues such as how model predictions might be affected by inflation do not impact our experimental design.

\begin{table*}[t]
\centering
\small
\begin{tabular}{p{0.12\linewidth}p{0.12\linewidth}p{0.08\linewidth}p{0.55\linewidth}}
\toprule
\textbf{Dataset} & \textbf{Feature} & \textbf{Type} & \textbf{Possible values} \\
\midrule
\multirow{1}{*}{Income} & age & Integer & $\{17, 18,\ldots, 95, 96\}$ \\
& education & Ordinal & \{`N/A - no schooling completed', `Nursery school / preschool', `Kindergarten', `1st grade only', `2nd grade', `3rd grade', `4th grade', `5th grade', `6th grade', `7th grade', `8th grade', `9th grade', `10th grade', `11th grade', `12th grade, no diploma', `Regular high school diploma', `GED or alternative credential', `Some college, less than 1 year', `Some college, 1 or more years, no degree', `Associate's degree', `Bachelor's degree', `Master's degree', `Professional degree beyond a bachelor's degree', `Doctorate degree'\} \\
\midrule
\multirow{1}{*}{House prices} & area & Ordinal & \{$500$, $1000$, $1500$, $2000$, $2500$, $3000$, $3500$, $4000$, $4500$, $5000$, $5500$, $6000$, $6500$, $7000$, $7500$, $8000$, $8500$, $9000$, $9500$, $10000$\} \\
& bedrooms & Integer & \{$1$, $2$, $3$, $4$, $5$\} \\
& bathrooms & Integer & \{$1$, $2$, $3$, $4$\}\\
& floors & Integer & \{$1$, $2$, $3$, $4$\}\\
\midrule
\multirow{1}{*}{Heart disease} & age & Ordinal & \{$30$, $35$, $40$, $45$, $50$, $55$, $60$, $65$, $70$, $75$, $80$\} \\
& sex & Categorical & \{`Female', `Male'\} \\
& systolic\_bp & Ordinal & \{$110$, $120$, $130$, $140$, $150$, $160$, $170$, $180$\} \\
& total\_cholesterol & Ordinal & \{$150$, $165$, $180$, $195$, $210$, $225$, $240$, $255$, $270$, $285$, $300$\} \\
\bottomrule
\end{tabular}
\caption{\textbf{Features in our three datasets.} The income dataset has two features. The house prices and heart disease dataset both have four features. All features are discrete to ensure that we can generate datasets that contain all feature combinations.}\label{tab:dataset_features}
\end{table*}

%%%%%%%%%%%%%%%%%%%%%%%%%%%%%%%%%%%%%%%%%%%%%%%%%%%%%%%%%%%%%%%%%%%%%%%%%%%%%%%%%%%%%%%%%%
\subsection{Models and inference environment}
%%%%%%%%%%%%%%%%%%%%%%%%%%%%%%%%%%%%%%%%%%%%%%%%%%%%%%%%%%%%%%%%%%%%%%%%%%%%%%%%%%%%%%%%%%
Open source models are downloaded from Hugging Face and run using the vLLM inference library \cite{kwon2023efficient}. Licenses are listed on the Hugging Face pages. These models use the Llama 3 Community License, the Gemma Terms of Use, and the MIT license (DeepSeek-R1 models). Locally run experiments were performed on four NVIDIA H100 GPUs and took approximately $20$ hours. Proprietary models are used through their respective APIs. Details of specific endpoints are provided in Table \ref{tab:models}.  

In the main experiment, we use temperature $0$ for all models except o3, which has a fixed temperature of $1$. Llama 3.3 70B is provided with the system prompt: `You are a helpful assistant.' Gemma models are not trained to accept system prompts and DeepSeek discourages the use of system prompts in the R1 family of models \cite{deepseekai2025deepseekr1distillllama70b}. The proprietary models are not given system prompts.

\begin{table*}[]
    \centering
    \small
    \begin{tabular}{lll}
        \toprule
        \textbf{Provider} & \textbf{Model} & \textbf{Hugging Face URL and API endpoints} \\
         \midrule
        Google & Gemma 2 27B & \href{https://huggingface.co/google/gemma-2-27b}{https://huggingface.co/google/gemma-2-27b} \\
        Meta & Llama 3.3 70B Instruct & \href{https://huggingface.co/meta-llama/Llama-3.3-70B-Instruct}{https://huggingface.co/meta-llama/Llama-3.3-70B-Instruct} \\
        DeepSeek & DeepSeek-R1 Qwen 32B & \href{https://huggingface.co/deepseek-ai/DeepSeek-R1-Distill-Qwen-32B}{https://huggingface.co/deepseek-ai/DeepSeek-R1-Distill-Qwen-32B} \\
        DeepSeek & DeepSeek-R1 Llama 70B & \href{https://huggingface.co/deepseek-ai/DeepSeek-R1-Distill-Llama-70B}{https://huggingface.co/deepseek-ai/DeepSeek-R1-Distill-Llama-70B} \\
        Anthropic & Claude Sonnet 3.7 &    claude-3-7-sonnet-20250219     \\
        OpenAI & GPT-4.1 &    gpt-4.1-2025-04-14     \\
        OpenAI & o3 &     o3-2025-04-16    \\
         \bottomrule
    \end{tabular}
    \caption{\textbf{Model details}. The majority of our inference was conducted using vLLM \cite{kwon2023efficient} with Hugging Face-downloaded models. Endpoints for proprietary models are also provided.} \label{tab:models}
\end{table*}

%%%%%%%%%%%%%%%%%%%%%%%%%%%%%%%%%%%%%%%%%%%%%%%%%%%%%%%%%%%%%%%%%%%%%%%%%%%%%%%%%%%%%%%%%%
\subsection{Prompting settings}\label{app:prompting_settings}
%%%%%%%%%%%%%%%%%%%%%%%%%%%%%%%%%%%%%%%%%%%%%%%%%%%%%%%%%%%%%%%%%%%%%%%%%%%%%%%%%%%%%%%%%%

\begin{figure*}[p]
  \centering

  \begin{minipage}{\textwidth}
    \centering 
    \includegraphics[width=\textwidth]{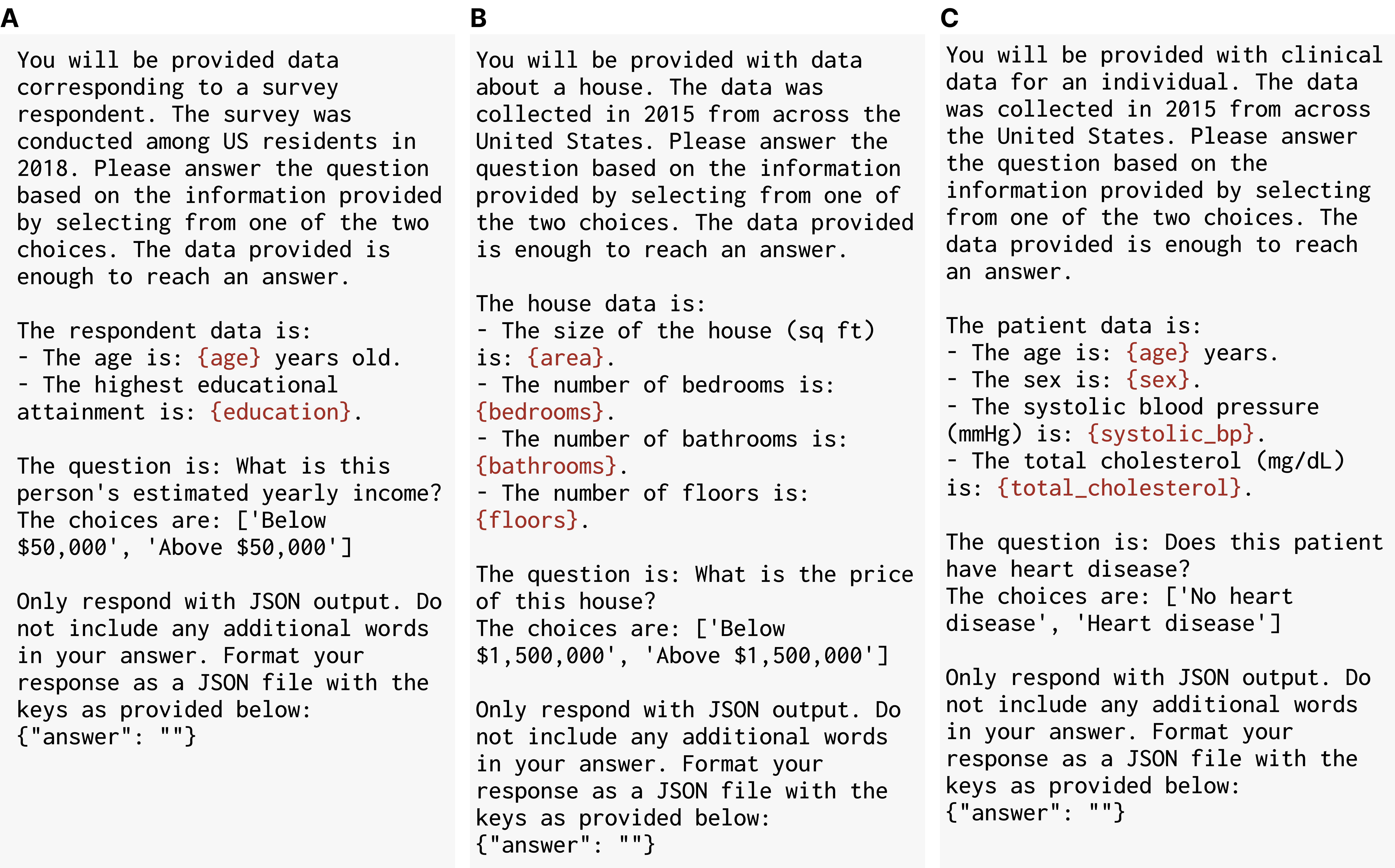}
        \caption{\textbf{Initial prompts to elicit model predictions for income (A), house prices (B), and heart disease (C) datasets.} These predictions are used as the first stage in the experimental pipeline. These prompts are templates where specific feature values are imputed based on the tabular data instance. \label{app_fig:prompts_1}}
  \end{minipage}\par\bigskip

  \begin{minipage}{\textwidth}
    \includegraphics[width=\textwidth]{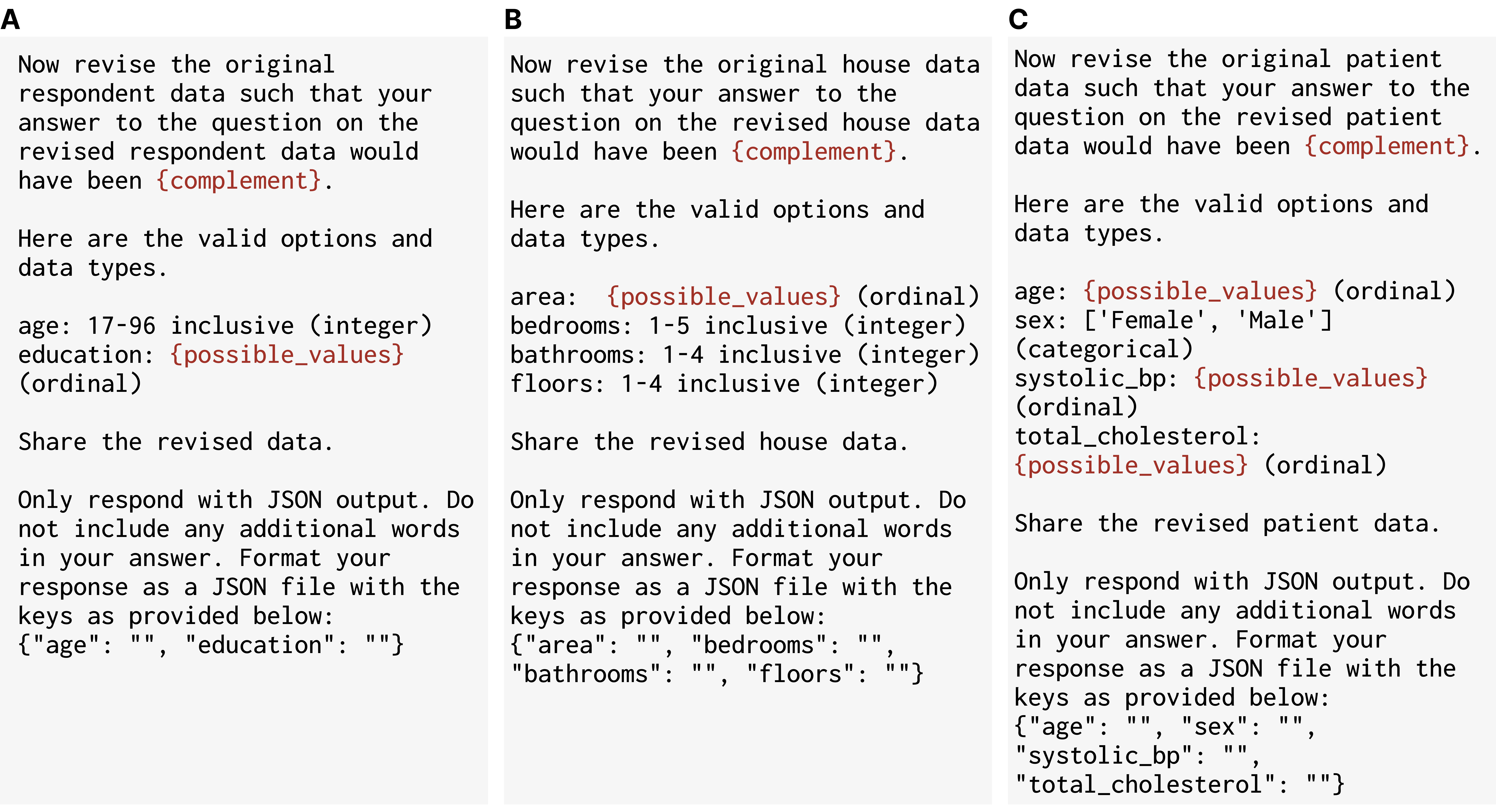}
        \caption{\textbf{Unconstrained prompts for income (A), house prices (B), and heart disease (C) datasets.} In our first experiment, we ask each model to provide a counterfactual input with no constraints on minimality. Below, we provide the prompts for each dataset. In all prompts, the value \texttt{\textcolor{BrickRed}{\{possible\_values\}}} refers to the complete list of ordinal values as provided in Table \ref{tab:dataset_features}, which we omit repeating for brevity. The value \texttt{\textcolor{BrickRed}{\{complement\}}} refers to the complement of the choice originally predicted by the model. See the prompts in Figure \ref{app_fig:prompts_1} for the list of choices. \label{app_fig:prompts_2}}
  \end{minipage}
\end{figure*}

\begin{figure*}[p]
  \centering
  \begin{minipage}{\textwidth}
    \centering 
    \includegraphics[width=\textwidth]{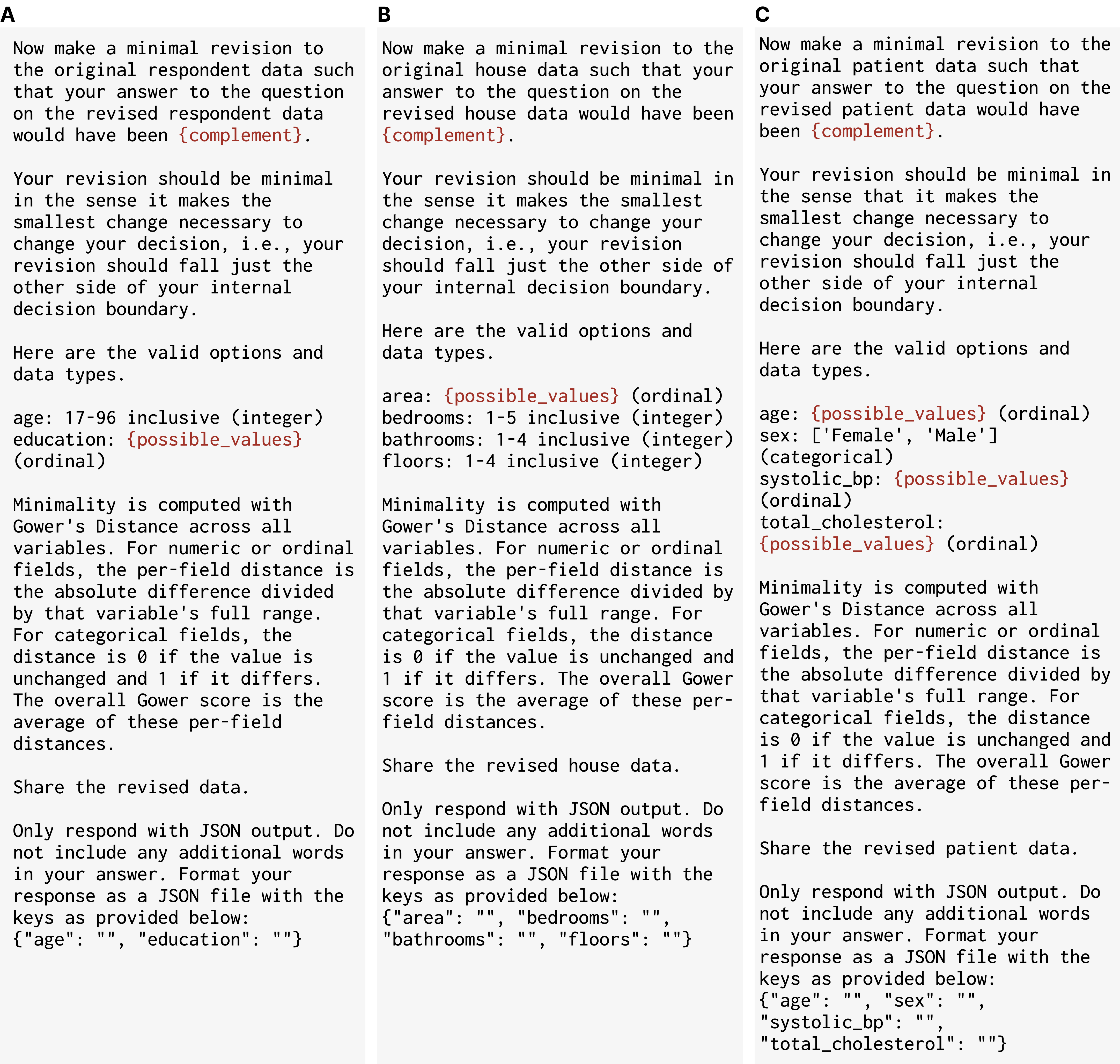}
        \caption{\textbf{Minimal prompts for income (A), house prices (B), and heart disease (C) datasets.} In our minimal prompt setting, we again ask each model to provide a counterfactual input, though we now instruct the model to provide the smallest change necessary to flip its decision. We also instruct the model to calculate minimality using Gower's Distance, and provide a brief explanation of Gower's Distance in each prompt. Below, we provide the prompts for each dataset. In all prompts, the value \texttt{\textcolor{BrickRed}{\{possible\_values\}}} refers to the complete list of ordinal values as provided in Table \ref{tab:dataset_features}, which we omit repeating for brevity. The value \texttt{\textcolor{BrickRed}{\{complement\}}} refers to the complement of the choice originally predicted by the model. See the prompts in Figure \ref{app_fig:prompts_1} for the list of choices. \label{app_fig:prompts_3}}
  \end{minipage}\par\bigskip
\end{figure*}

Our experimental pipeline has two steps. First, iterate through every instance in the dataset to collect the models' predictions. Then, we ask the model to provide an SCE that would flip its prediction. An explicit verification step is not required because we can use the prediction from the first stage as a lookup table to see what the model would have predicted for the counterfactual input. The prediction-eliciting prompts are shown in Figure \ref{app_fig:prompts_1}. The SCE-eliciting prompts are shown in Figure \ref{app_fig:prompts_2} and Figure \ref{app_fig:prompts_3} for the unconstrained and minimal prompting settings, respectively. 

%%%%%%%%%%%%%%%%%%%%%%%%%%%%%%%%%%%%%%%%%%%%%%%%%%%%%%%%%%%%%%%%%%%%%%%%%%%%%%%%%%%%%%%%%%
\subsection{Evaluation metrics}\label{app:gowers_distance}
%%%%%%%%%%%%%%%%%%%%%%%%%%%%%%%%%%%%%%%%%%%%%%%%%%%%%%%%%%%%%%%%%%%%%%%%%%%%%%%%%%%%%%%%%%

We consider three evaluation metrics: validity, excess distance, and exact match. For excess distance, our main minimality metric, we use \textit{Gower's Distance} as the distance function. In the general case, Gower's Distance is defined between $p$-dimensional items $z_i, z_j\in \mathcal{Z}$ as
\[
    d(z_i, z_j) = \frac{\sum_{k=1}^p w_{ijk} s_{ijk}}{\sum_{k=1}^p w_{ijk}},
\]
where $w_{ijk}$ are non-negative weights, typically set to $1$ if considering all feature comparisons equally. If the $k$-th variable is binary, $s_{ijk}$ is defined
\[
    s_{ijk} = \mathbb{I}[z_{ik} = z_{jk}].
\]
If the variable is continuous, then
\[
    s_{ijk} = 1 - \frac{|z_{ik} - z_{jk}|}{\max_{z_l\in\mathcal{Z}} z_{lk} - \min_{z_l\in \mathcal{Z}}z_{lk}}.
\]
If the variable is ordinal, then, similarly,
\[
    s_{ijk} = 1 - \frac{|r_i - r_j|}{\max\{r\} - \min\{ r\}}
\]
with $r$ being the ranks corresponding to the ordered categories of the $k$-th variable~\cite{podaniExtendingGowersGeneral1999}.

Given all variables in our datasets are numerical, binary or ordinal (converted to ranks, then treated as numerical) Gower's Distance simplifies to
\[
d(z_i, z_j) = \frac{1}{p}\sum_{k=1}^{p}\dfrac{|z_{ik} - z_{jk}|}{\max(z_{k}) - \min(z_{k})}.
\]
Intuitively, this metric represents the mean of the per-feature distances, where each per-feature distance is the difference between the item values, normalised by the feature's range. The resulting metric ranges from $0$ (identical instances) to $1$ (maximally different). The distance between two dataset instances can be interpreted as a fraction of the maximum distance between any two dataset instances.

%%%%%%%%%%%%%%%%%%%%%%%%%%%%%%%%%%%%%%%%%%%%%%%%%%%%%%%%%%%%%%%%%%%%%%%%%%%%%%%%%%%%%%%%%%
\section{Complete results for the main experiment}
%%%%%%%%%%%%%%%%%%%%%%%%%%%%%%%%%%%%%%%%%%%%%%%%%%%%%%%%%%%%%%%%%%%%%%%%%%%%%%%%%%%%%%%%%%

\subsection{Full result tables}\label{app:full_results}

Table~\ref{tab:full_results} shows the full results for the main experiment. We present the average validity and excess distance ($\mathrm{ED}$).

\begin{table*}[ht]
\centering
\small
\begin{tabular}{l|l|cc|cc}
\toprule
\multirow{2}{*}{\textbf{Dataset}} & \multirow{2}{*}{\textbf{Model}}
& \multicolumn{2}{c|}{\textbf{Unconstrained prompting}}
& \multicolumn{2}{c}{\textbf{Minimal prompting}} \\[2pt]
& & \textbf{Val} & $\mathbf{ED}$ & \textbf{Val} & $\mathbf{ED}$ \\
\midrule
\multirow{6}{*}{Income}
& Gemma 2 27B       & $99.48$ &  $0.1504$ & $49.95$ &$0.0151$\\
& Llama 3.3 70B     & $100.0$ & $0.1393$ & $15.21$ & $0.0305$ \\
& DeepSeek-R1 32B   & $88.99$ & $0.1116$ & $10.52$ & $0.0111$ \\
& DeepSeek-R1 70B   & $92.50$ & $0.0603$ & $10.26$ & $0.0116$ \\
& Claude Sonnet 3.7 & $99.53$ & $0.1134$ & $67.08$ & $0.0203$ \\
& GPT-4.1 & $99.47$ & $0.1505$ & $51.51$ & $0.0256$ \\
& o3 & $100.0$ & $0.1629$ & $49.58$ & $0.0177$ \\
\midrule
\multirow{6}{*}{House prices}
& Gemma 2 27B       & $100.0$ & $0.3158$ & $14.81$ & $0.0126$ \\
& Llama 3.3 70B     & $100.0$ & $0.4092$ & $10.31$ & $0.0099$ \\
& DeepSeek-R1 32B   &  $91.94$ & $0.3522$ & $12.78$ & $0.0063$ \\
& DeepSeek-R1 70B   &  $77.50$ & $0.2981$ & $21.44$ & $0.0254$ \\
& Claude Sonnet 3.7 & $99.88$ & $0.3739$ & $18.38$ & $0.0689$ \\
& GPT-4.1 & $100.0$ & $0.3760$ & $19.31$ & $0.0470$ \\
& o3 & $100.0$ & $0.4188$ & $40.34$ & $0.0145$ \\

\midrule
\multirow{6}{*}{Heart disease}
& Gemma 2 27B       & $100.0$ & $0.1913$ & $31.57$ & $0.0104$ \\
& Llama 3.3 70B     & $100.0$ & $0.2951$ & $25.26$ & $0.0114$ \\
& DeepSeek-R1 32B   &  $93.68$ & $0.4073$ & $33.69$ & $0.0173$ \\
& DeepSeek-R1 70B   &  $91.27$ & $0.2953$ & $25.46$ & $0.0073$ \\
& Claude Sonnet 3.7 & $100.0$ & $0.4723$ & $23.09$ & $0.0714$ \\
& GPT-4.1 & $100.0$ & $0.3415$ & $19.21$ & $0.0050$ \\
& o3 & $100.0$ & $0.4985$ & $31.75$ & $0.0210$ \\

\bottomrule
\end{tabular}
\caption{\textbf{The properties of SCEs at temperature $\mathbf{0.0}$}.  
Validity is the percentage of times an SCE leads to the target prediction when evaluated using a new instance of the model. $\mathrm{ED}$ is the excess distance, as defined in Section~\ref{sec:metrics}.}
\label{tab:full_results}
\end{table*}

\subsection{Density heatmaps}\label{app:density_plots}

\begin{figure*}[h]
    \centering 
    \includegraphics[width=1\textwidth]{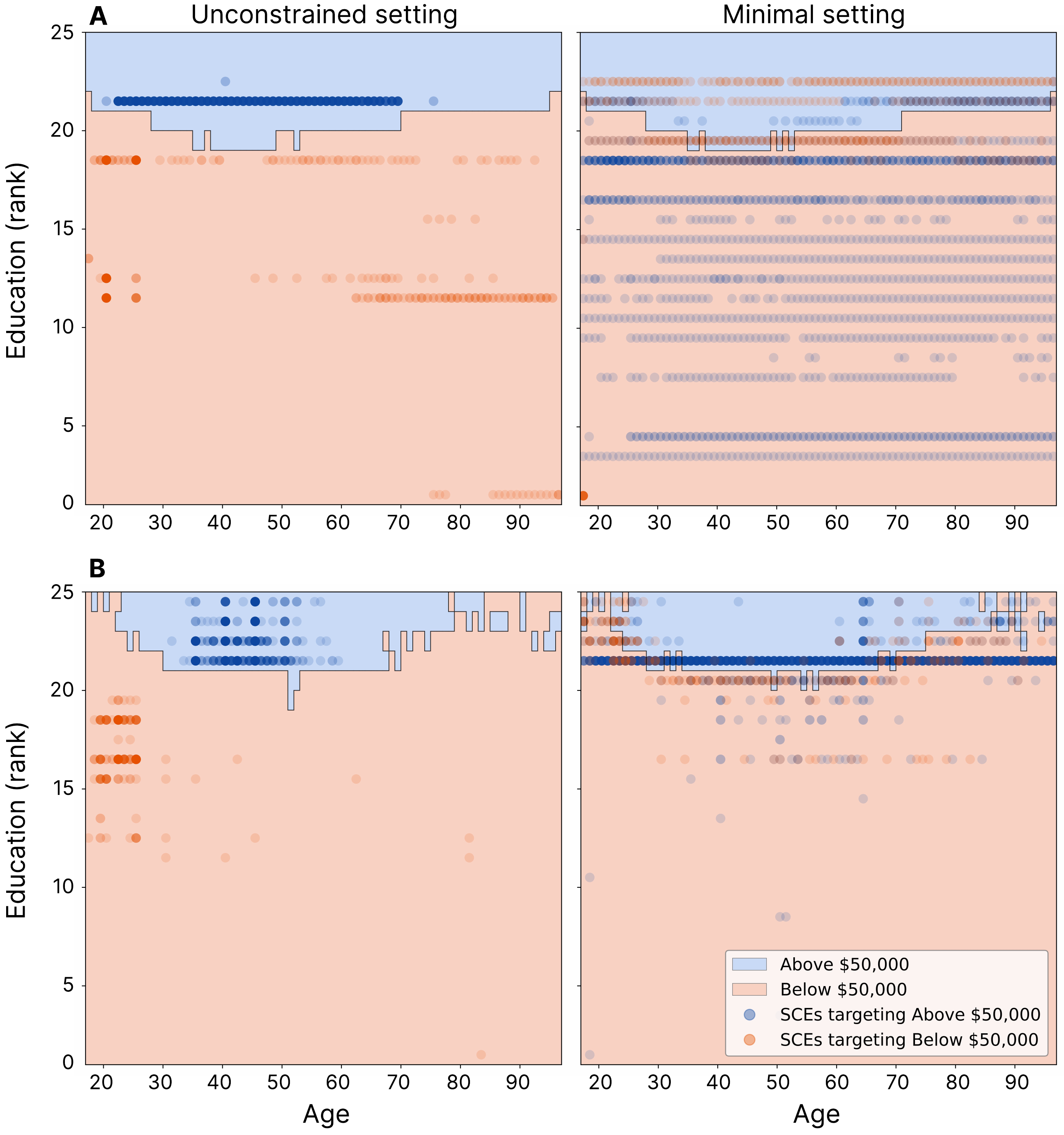}
        \caption{\textbf{The distribution of SCEs for Llama 3.3 70B (A) and o3 (B).} This shows the full distribution of SCEs for every starting point in the income dataset. \label{fig:density_plots}}
\end{figure*}

Figure \ref{fig:density_plots} shows the distribution of SCEs for Llama 3.3 70B and o3 across all instances in the income dataset. In the unconstrained prompting setting, the SCEs cluster around distinct regions for both models. In the minimal prompting setting, the SCEs often fall short of crossing the decision boundary. o3 is visibly better than Llama 3.3 70B at generating minimal SCEs, with more of the SCEs closer to the decision boundary. However, there are still many invalid SCEs.

%%%%%%%%%%%%%%%%%%%%%%%%%%%%%%%%%%%%%%%%%%%%%%%%%%%%%%%%%%%%%%%%%%%%%%%%%%%%%%%%%%%%%%%%%%
\section{Robustness analysis supplementary information}
%%%%%%%%%%%%%%%%%%%%%%%%%%%%%%%%%%%%%%%%%%%%%%%%%%%%%%%%%%%%%%%%%%%%%%%%%%%%%%%%%%%%%%%%%%

\subsection{Distance function sensitivity}\label{app:distance_function_sensitivity}

In Section \ref{sec:minimal_prompting_results}, we report Llama 3.3 70B's results under different distance functions. The exact definitions of each metric are reported below.

\paragraph{$L_1$} This is defined as in~\citet{wachter2017counterfactual}, where raw $L_1$ distance is normalised across each feature $k$ using the median absolute deviation,
\[
    \mathrm{MAD}_k = \mathrm{median}_{z_i\in\mathcal{Z}}(z_{ik} - \mathrm{median}_{z_j\in\mathcal{Z}}(z_{jk}))
\]
Our distance metric is thus defined
\[
    d_{L1}(z_i, z_j) = \sum_{k=1}^p \frac{\lvert z_{ik} - z_{jk}\rvert}{\mathrm{MAD}_k}.
\]

\paragraph{$L_2$} Again, following~\citet{wachter2017counterfactual}, we normalise using the standard deviation over each feature $k$. Our distance metric is defined
\[
    d_{L_2}(z_i, z_j) = \sum_{k = 1}^p \frac{(z_{ik} - z_{jk})^2}{\sigma_{z_l\in\mathcal{Z}}(z_{lk})}.
\]

\paragraph{Cosine} To capture semantic differences between inputs, we consider an embedding-based distance. We first contextualise the tabular by inserting it into the templates in Figure \ref{app_fig:prompts_1}. Then, we extract the respondent data part of the template (i.e. just the list of features and their values) and use the \texttt{all-mpnet-base-v2} embedding model, a fine-tuned variation of Microsoft's MPNet model~\cite{song2020mpnet}, to extract embeddings $\mathrm{enc}(x_i)\in\mathbb{R}^{768}$. Our distance function is $1$ minus the cosine similarity of the two embeddings
\[
    d_\mathrm{cosine}(z_i, z_j) = 1-S_C(\mathrm{enc}(\phi(z_i)), \mathrm{enc}(\phi(z_j))).
\]
Subtracting the cosine similarity from $1$ ensures that it is a distance metric rather than a similarity metric.

%%%%%%%%%%%%%%%%%%%%%%%%%%%%%%%%%%%%%%%%%%%%%%%%%%%%%%%%%%%%%%%%%%%%%%%%%%%%%%%%%%%%%%%%%%
\subsection{Prompt sensitivity}\label{app:prompt_sensitivity}
\begin{figure*}[h]
    \centering 
    \includegraphics[width=1\textwidth]{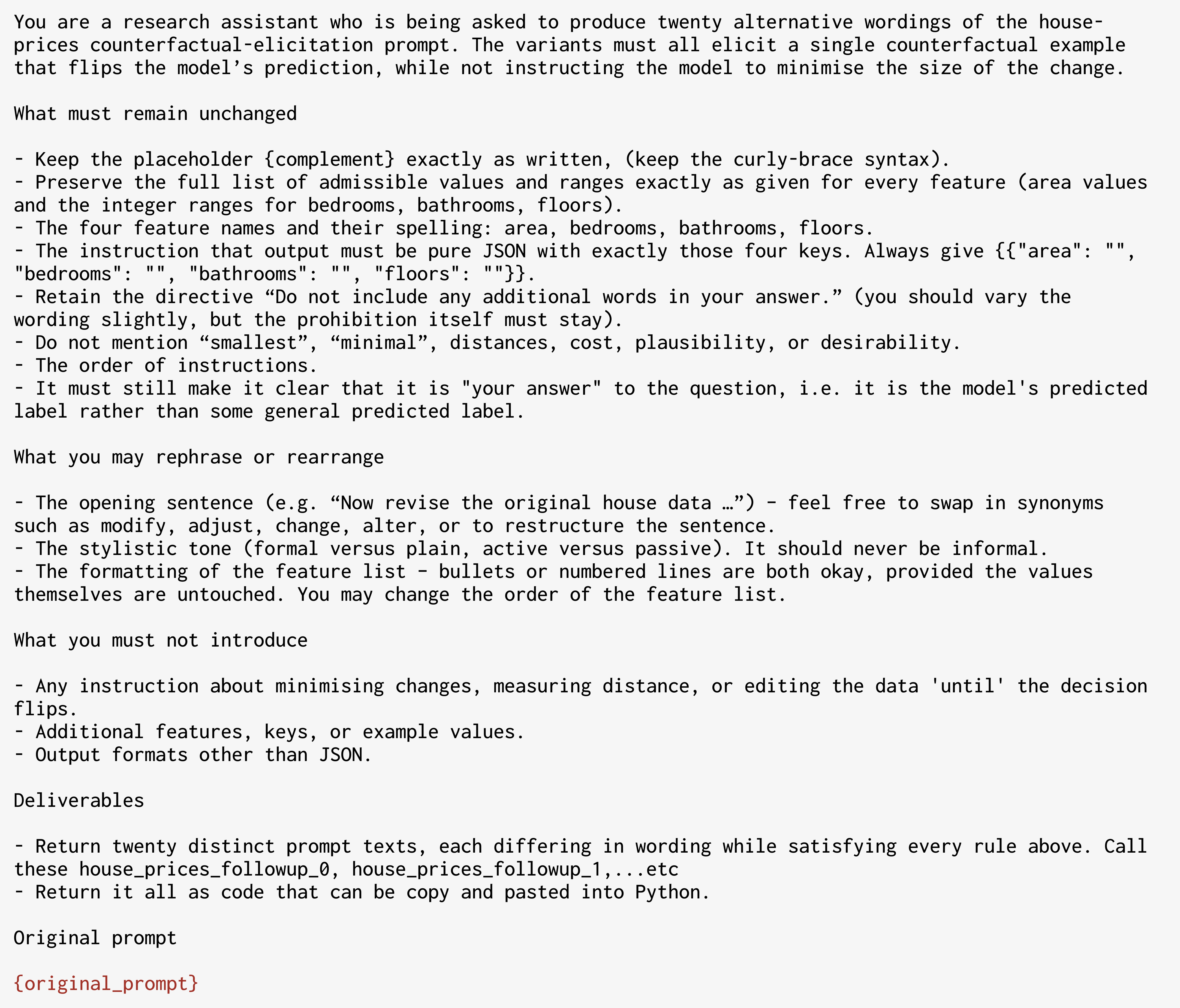}
        \caption{\textbf{Prompt given to o3 to generate the prompt perturbations.} The value \texttt{\textcolor{BrickRed}{\{original\_prompt\}}} shows where the original prompt is inserted into the template. \label{fig:prompt_sensitivity_app}}
\end{figure*}

%%%%%%%%%%%%%%%%%%%%%%%%%%%%%%%%%%%%%%%%%%%%%%%%%%%%%%%%%%%%%%%%%%%%%%%%%%%%%%%%%%%%%%%%%%
Figure \ref{fig:prompt_sensitivity_app} shows the prompt that we gave to OpenAI's o3 to generate the prompt perturbations for the experiment in Section \ref{sec:prompt_sensitivity}. We generated 20 versions of the SCE-eliciting prompt and evaluated how Llama 3.3 70B's behaviour changed.

%%%%%%%%%%%%%%%%%%%%%%%%%%%%%%%%%%%%%%%%%%%%%%%%%%%%%%%%%%%%%%%%%%%%%%%%%%%%%%%%%%%%%%%%%%
\subsection{Results at temperature $\mathbf{1.0}$}\label{app:temp_1}
%%%%%%%%%%%%%%%%%%%%%%%%%%%%%%%%%%%%%%%%%%%%%%%%%%%%%%%%%%%%%%%%%%%%%%%%%%%%%%%%%%%%%%%%%%

\begin{figure*}[p]
  \centering

  \begin{minipage}{\textwidth}
    \centering 
    \includegraphics[width=1\textwidth]{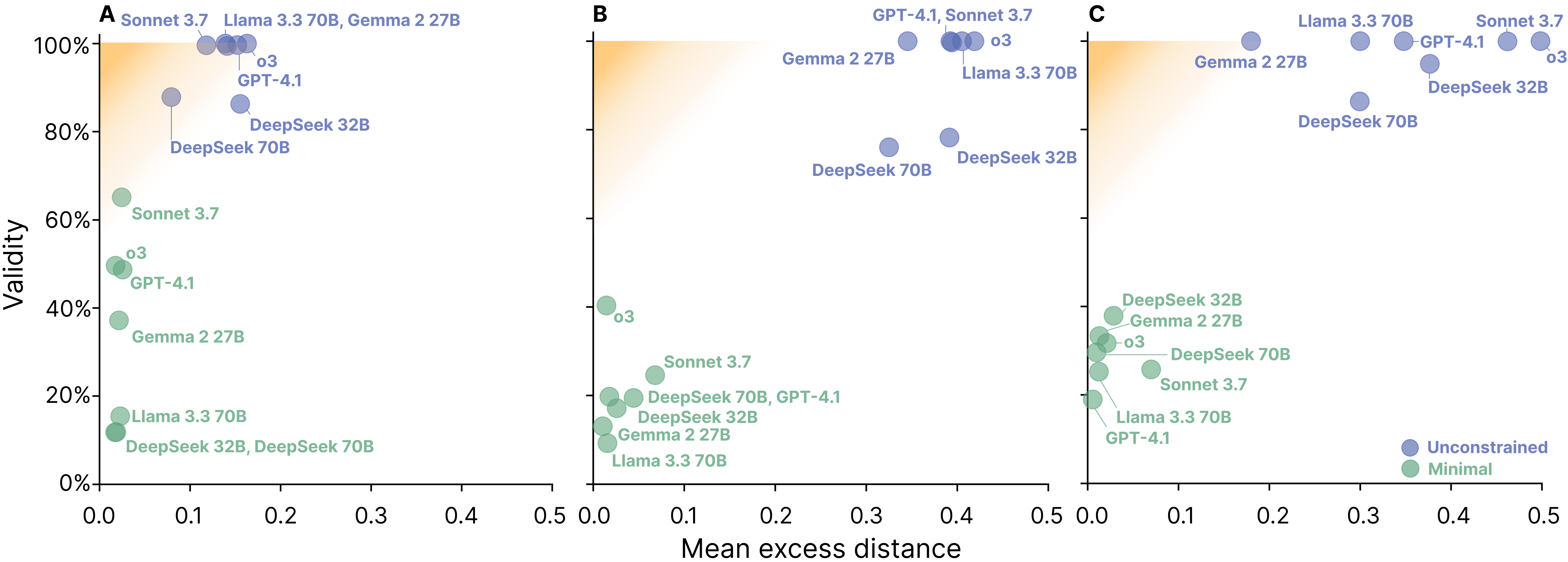}
        \caption{\textbf{SCE validity and minimality in the income (A), house prices (B), and heart disease (C) datasets at temperature $\mathbf{1.0}$.} In the unconstrained prompting setting, models are able to provide valid SCEs, however, they are far from minimal. In the minimal prompting setting, validity drops sharply. No model can satisfy both criteria. Orange regions indicate the direction of increasing validity and minimality. \label{fig:scatter_plot_temp1}}
  \end{minipage}\par\bigskip

  \begin{minipage}{\textwidth}
      \centering
      \small
      \begin{tabular}{l|l|cc|cc}
      \toprule
      \multirow{2}{*}{\textbf{Dataset}} & \multirow{2}{*}{\textbf{Model}}
      & \multicolumn{2}{c|}{\textbf{Unconstrained prompting}}
      & \multicolumn{2}{c}{\textbf{Minimal prompting}} \\[2pt]
      & & \textbf{Val}& $\mathbf{ED}$
          & \textbf{Val} & $\mathbf{ED}$ \\
      \midrule
      \multirow{7}{*}{Income}
      & Gemma 2 27B       &  $99.43$ & $0.1409$ & $37.2$ & $0.0215$ \\
      & Llama 3.3 70B     & $100.0$ & $0.1389$ & $15.5$ & $0.0229$ \\
      & DeepSeek-R1 32B   &  $86.34$ & $0.1553$ & $11.8$ & $0.0184$ \\
      & DeepSeek-R1 70B   &  $87.85$ & $0.0793$ & $11.8$ & $0.0171$ \\
      & Claude Sonnet 3.7 &  $99.63$ & $0.1179$ & $65.1$ & $0.0244$ \\
      & GPT-4.1           &  $99.69$ & $0.1520$ & $48.8$ & $0.0255$ \\
      & o3                & $100.0$ & $0.1629$ & $49.6$ & $0.0177$ \\
      \midrule
      \multirow{7}{*}{House prices}
      & Gemma 2 27B       & $100.0$ & $0.3454$ & $13.1$ & $0.0103$ \\
      & Llama 3.3 70B     & $100.0$ & $0.4051$ &  $9.25$ & $0.0155$ \\
      & DeepSeek-R1 32B   &  $78.25$ & $0.3913$ & $17.2$ & $0.0257$ \\
      & DeepSeek-R1 70B   &  $76.07$ & $0.3250$ & $19.8$ & $0.0176$ \\
      & Claude Sonnet 3.7 &  $99.75$ & $0.3939$ & $24.6$ & $0.0679$ \\
      & GPT-4.1           & $100.0$ & $0.3926$ & $19.5$ & $0.0444$ \\
      & o3                & $100.0$ & $0.4188$ & $40.3$ & $0.0145$ \\
      \midrule
      \multirow{7}{*}{Heart disease}
      & Gemma 2 27B       & $100.0$ & $0.1797$ & $33.3$ & $0.0129$ \\
      & Llama 3.3 70B     & $100.0$ & $0.2999$ & $25.3$ & $0.0122$ \\
      & DeepSeek-R1 32B   &  $94.88$ & $0.3766$ & $37.9$ & $0.0283$ \\
      & DeepSeek-R1 70B   &  $86.37$ & $0.2994$ & $29.6$ & $0.0096$ \\
      & Claude Sonnet 3.7 &  $99.95$ & $0.4619$ & $25.8$ & $0.0699$ \\
      & GPT-4.1           & $100.0$ & $0.3478$ & $19.0$ & $0.0055$ \\
      & o3                & $100.0$ & $0.4985$ & $31.7$ & $0.0210$ \\
      \bottomrule
      \end{tabular}
      \captionof{table}{\textbf{Properties of self-generated counterfactual explanations under temperature $\mathbf{1.0}$}.  
      Validity is the percentage of times an SCE leads to the target prediction when evaluated on a new instance of the model. $\mathrm{ED}$ is the excess distance, as defined in Section~\ref{sec:metrics}.}
      \label{tab:full_results_temp1}
    \end{minipage}
    
\end{figure*}

Figure \ref{fig:scatter_plot_temp1} and Table \ref{tab:full_results_temp1} show the full results under temperature $1.0$. We find minimal difference in aggregate results. This is in line with a previous experiment by~\citet{dehghanighobadi2025llmsexplaincounterfactually}.

%%%%%%%%%%%%%%%%%%%%%%%%%%%%%%%%%%%%%%%%%%%%%%%%%%%%%%%%%%%%%%%%%%%%%%%%%%%%%%%%%%%%%%%%%%
\section{The limiting factors of model performance}
%%%%%%%%%%%%%%%%%%%%%%%%%%%%%%%%%%%%%%%%%%%%%%%%%%%%%%%%%%%%%%%%%%%%%%%%%%%%%%%%%%%%%%%%%%

\subsection{Decision boundary consistency}\label{app:prompt_perturbations}

\begin{figure*}[h]
    \centering 
    \includegraphics[width=1\textwidth]{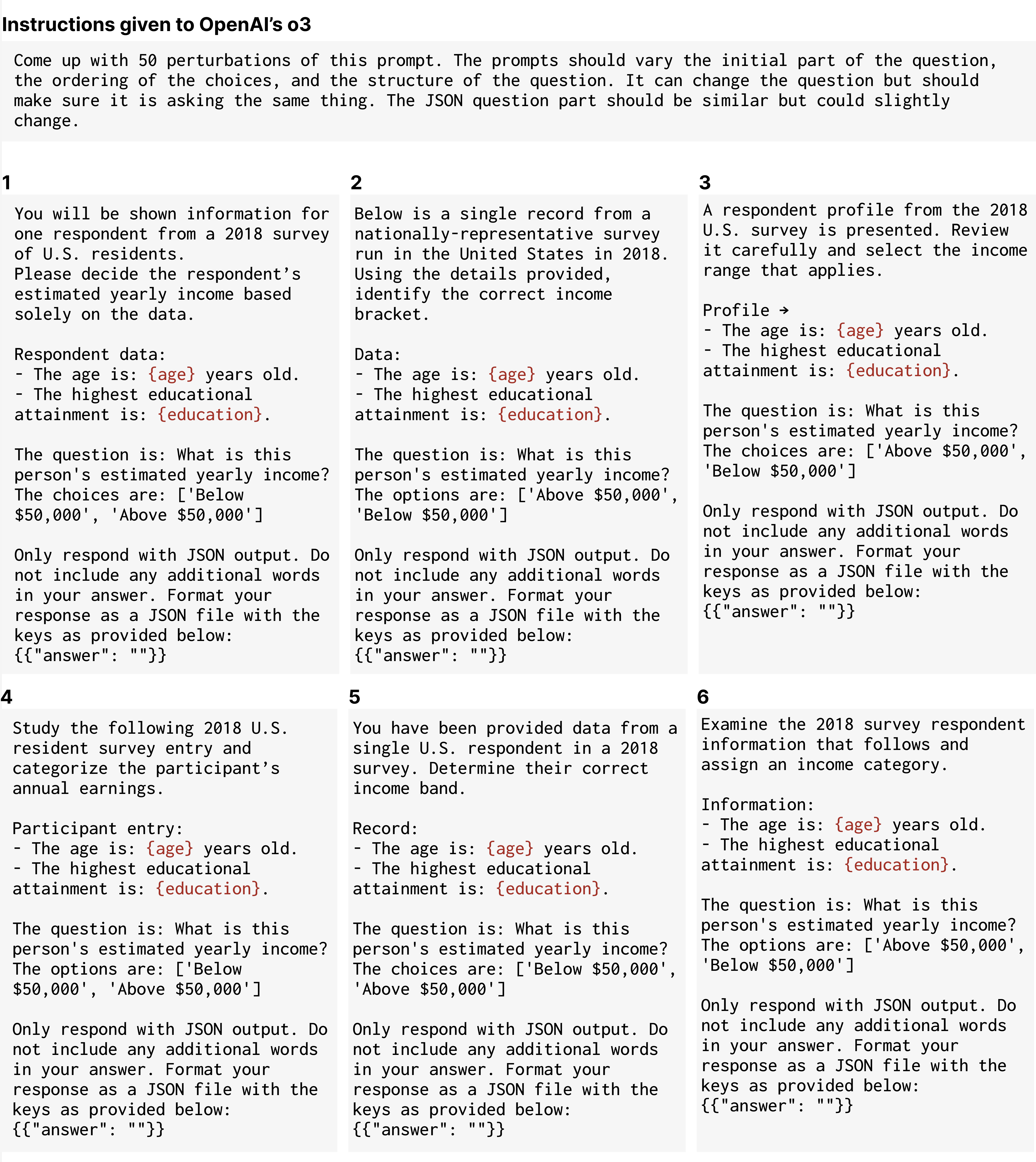}
        \caption{\textbf{Instructions passed to OpenAI's o3 model and six of the resulting prompt perturbations.} The values \texttt{\textcolor{BrickRed}{\{in\_red\}}} indicate where the numerical values are inserted into the templates. \label{fig:prompt_5}}
\end{figure*}

We considered how consistent Llama 3.3 70B's decision boundary was to prompt perturbations. A potential failure model for generating SCEs is that there is no consistent internal decision boundary; hence the notion of validity is ill-defined. To assess this, we generated $50$ perturbations of the income prediction task using OpenAI's o3 \cite{OpenAI2025o3o4mini}. The instruction passed to OpenAI's o3 model and six of the resulting perturbations of the prompt are shown in Figure \ref{fig:prompt_5}.

%%%%%%%%%%%%%%%%%%%%%%%%%%%%%%%%%%%%%%%%%%%%%%%%%%%%%%%%%%%%%%%%%%%%%%%%%%%%%%%%%%%%%%%%%%
\subsection{Operationalising distance}\label{app_understanding_gower}
%%%%%%%%%%%%%%%%%%%%%%%%%%%%%%%%%%%%%%%%%%%%%%%%%%%%%%%%%%%%%%%%%%%%%%%%%%%%%%%%%%%%%%%%%%

\begin{figure*}[h]
    \centering 
    \includegraphics[width=1\textwidth]{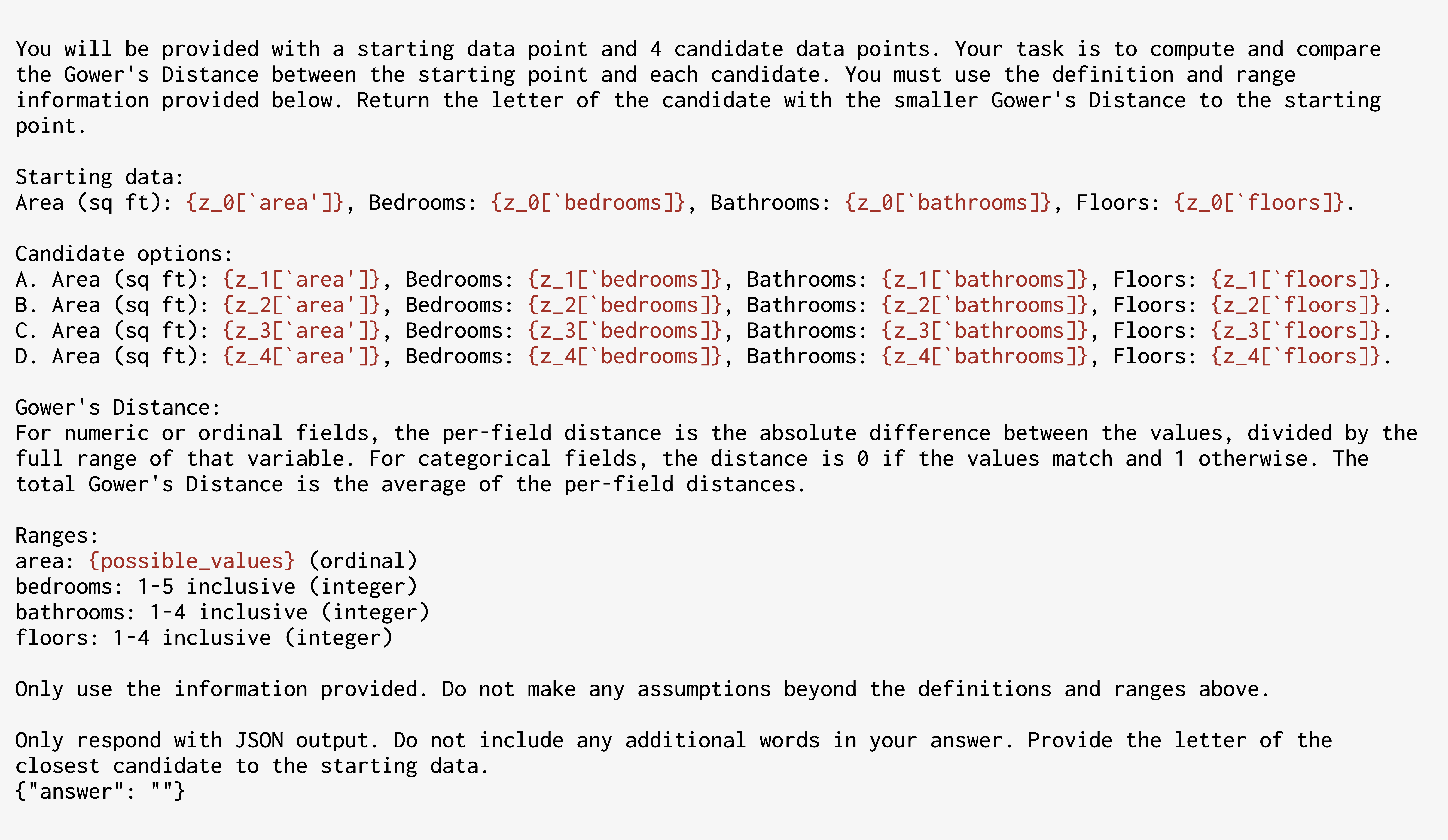}
        \caption{\textbf{Prompt for the operationalising distance experiment.} The values \texttt{\textcolor{BrickRed}{\{in\_red\}}} indicate where the sampled numerical values are inserted into the template. \label{fig:prompt_7}}
\end{figure*}

One reason why models may fail to provide valid and minimal SCEs is that they simply do not understand Gower's Distance. To test this, we design a baseline validation experiment using a multiple choice task. For each trial in our test, we provide a model with a starting point from the house price dataset $z_0$ and $4$ randomly sampled alternative points $(z_1, z_2, z_3, z_4)$. The model's task is to identify the closest point to $z_0$ under Gower’s Distance. For each problem, we format the input using the template shown in Figure \ref{fig:prompt_7}.

We measure the exact match accuracy over $1000$ trials, where we use the implementation of Gower's Distance in Section \ref{app:gowers_distance} to compute the correct answer for comparison. Models are evaluated at temperature $0$ (except o3 at temperature $1.0$). We report the accuracies in Table~\ref{tab:distance_experiment}. Random accuracy is $25\%$.

\subsection{Self-prediction}\label{app:reasoning_traces}

\begin{figure*}
    \centering
    \includegraphics[width=\linewidth]{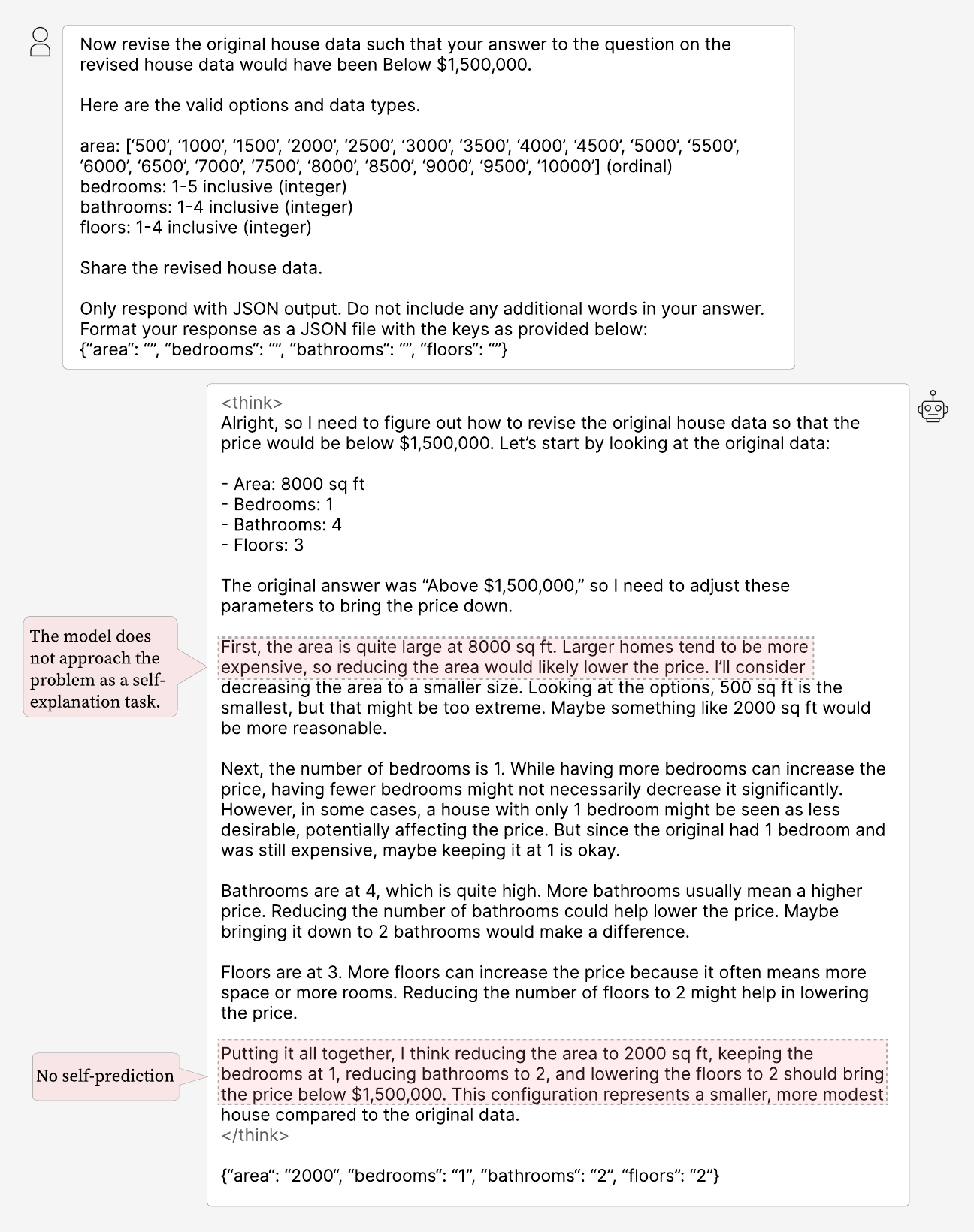}
    \caption{\textbf{A full reasoning trace from DeepSeek-R1 70B}. Excerpts from the model response suggest that, despite the question being clearly framed as a self-explanation task (`revise the original house data such that \textit{your} answer to the question\dots'), DeepSeek-R1 70B does not interpret the problem as such. The model instead appeals to general intuitions (e.g., `More bathrooms usually mean a higher price') and never considers its own decision-making process or explicitly self-predicts.}\label{fig:full_reasoning_trace}
\end{figure*}

Figure~\ref{fig:full_reasoning_trace} shows an example of a full reasoning trace from DeepSeek-R1 Llama 70B when attempting to generate a minimal SCE on the house price dataset. The model appears to misunderstand the \textit{self-}explanation elements of the task and does not self-predict in its chain-of-thought.

\section{Incentivising models to self-predict}\label{app:self_prediction_ablation}

We also conducted an ablation experiment where we edited the minimality prompt to explicitly encourage self-prediction. Previous studies have shown that prompt engineering based on human metacognition strategies can improve performance in metacognition-related tasks such as calibration between intrinsic uncertainty and stated uncertainty \cite{liu2025metafaith}. 

\begin{figure*}[h]
    \centering 
    \includegraphics[width=1\textwidth]{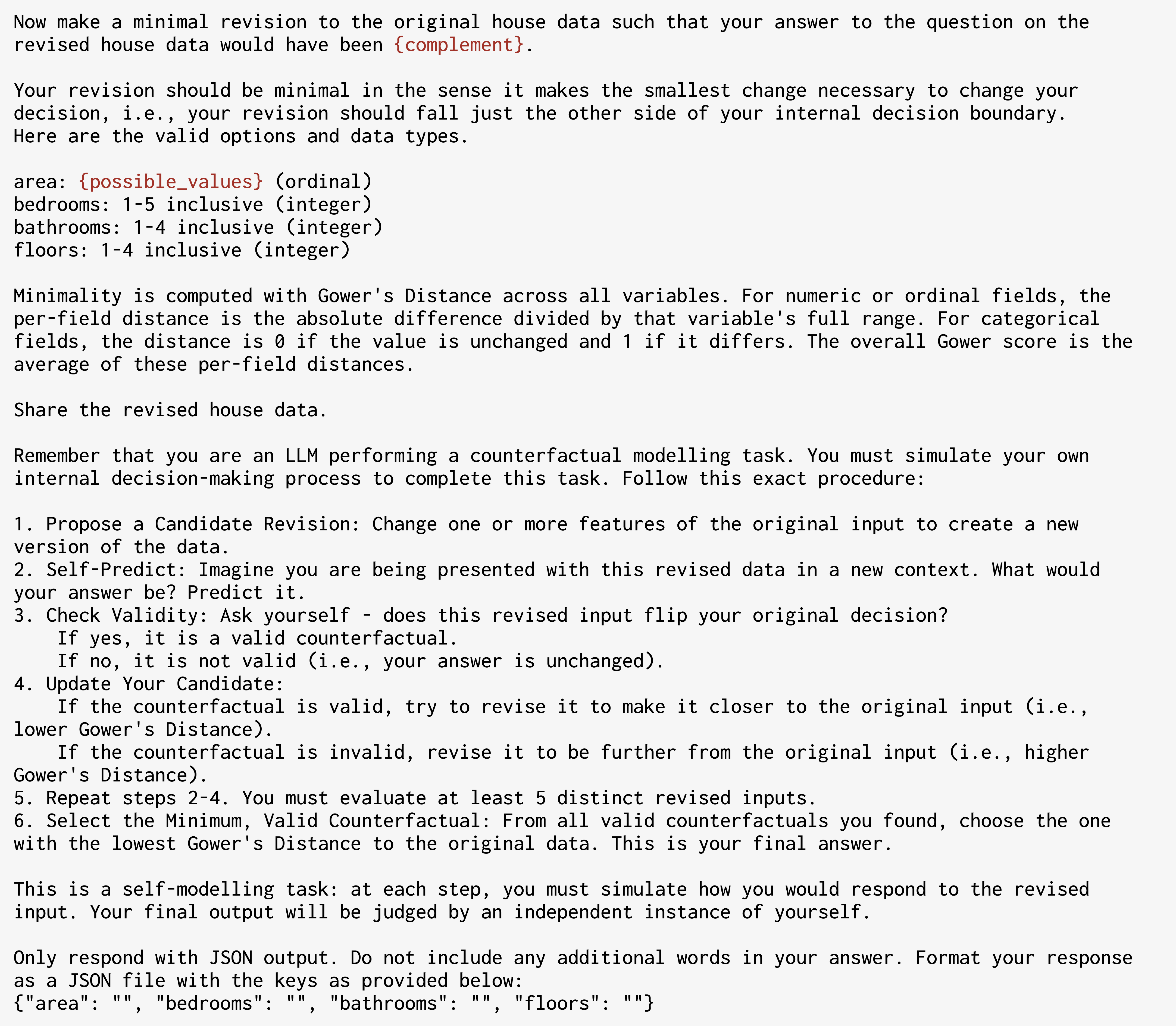}
        \caption{\textbf{Prompt for self-prediction ablation experiment.} We provide the models with a metacognition-inspired prompt with a six-step plan to generate a minimal valid counterfactual. This prompt encourages the LLMs to make explicit self-predictions for every candidate SCE they consider.\label{fig:prompt_4}}
\end{figure*}

To achieve this, we designed a six-step reasoning plan for the LLMs to follow (see Figure \ref{fig:prompt_4}). First, they must propose a candidate counterfactual. Second, they should make an explicit self-prediction by imagining that they are presented with the revised data in a new context. Third, they should assess the validity and minimality of the candidate SCE based on their self-prediction. Fourth, based on this assessment, they should update their candidate SCE. Fifth, they need to repeat the preceding steps for at least five distinct candidate counterfactuals. Sixth, they should select the minimum valid counterfactuals from all candidates considered. In addition, we tell the model that this is a \textit{self-modelling} task and the output will be evaluated by an independent instance of the model itself. 

We evaluate how DeepSeek-R1 70B performs on the house price dataset. First, the model follows the six steps well. We see the model systematically consider candidate SCEs and find it grappling with the \textit{self-}explanation criteria of the task to predict its own behaviour. In particular, we see phrases such as \textit{`Based on my internal model, this revised house would likely be priced below $\$1,500,000$.'}. In most cases, however, the model still makes general predictions without explicitly recognising the validity will be assessed by itself, e.g. \textit{Given that the area has doubled, it's likely that the price would increase substantially. However, I'm not sure.}

\begin{figure*}
    \centering
    \includegraphics[width=0.8\linewidth]{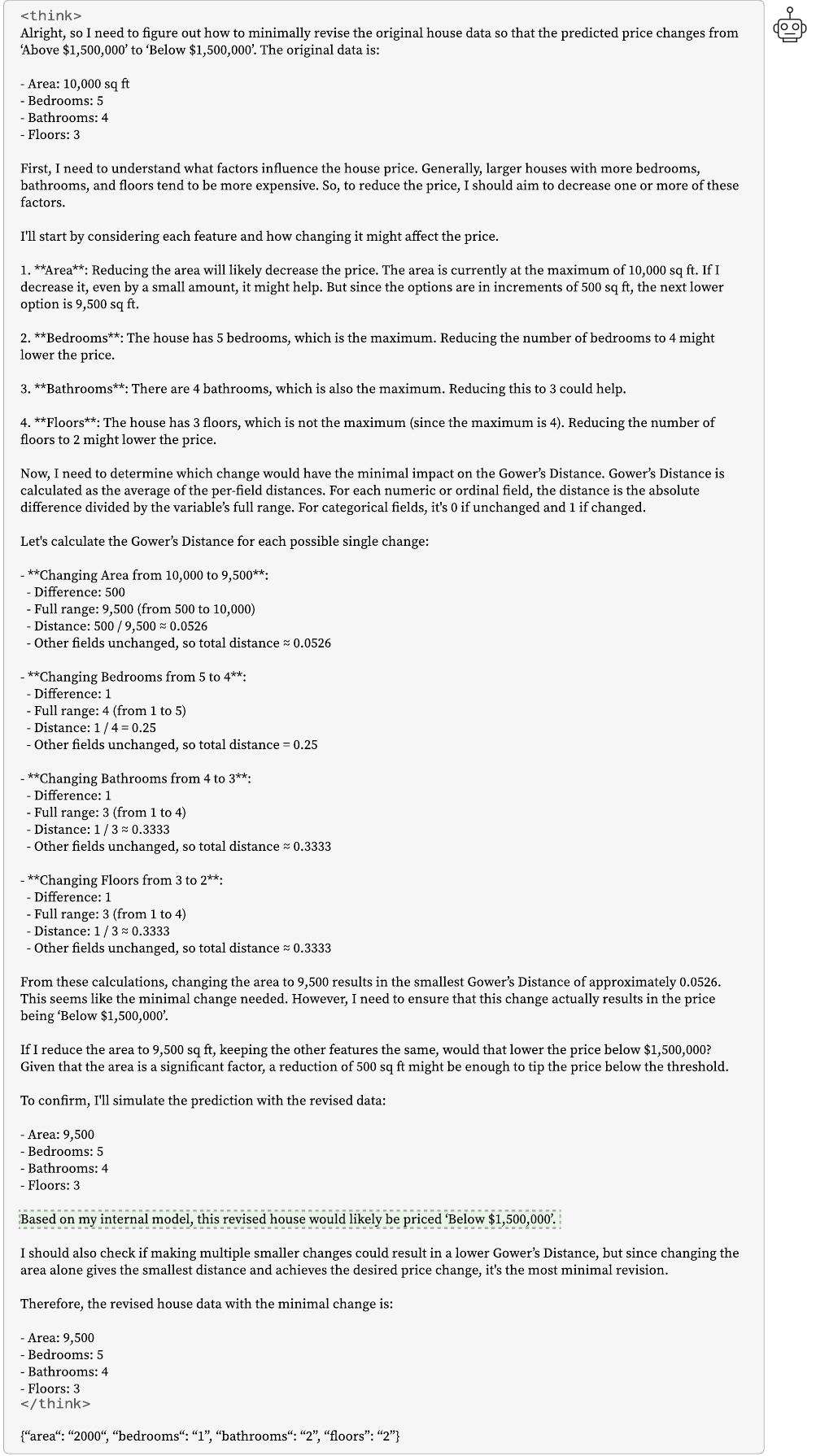}
    \caption{\textbf{A reasoning trace from DeepSeek-R1 70B following step-by-step self-prediction instructions}. We prompt the model using the template from~\S\ref{app:self_prediction_ablation}.}
    \label{fig:explicit_reasoning_trace}
\end{figure*}

Despite the model making self-predictions in-context, we find that the aggregate results are largely unchanged. Validity increases from $21.44\%$ to $23.34\%$, at the cost of excess distance increasing from $0.025$ to $0.033$. Anecdotally, we also find that self-predictions made in the reasoning traces are often wrong. Whether this is a fundamental limitation of LLMs is unclear and an interesting question for future work to address. We show an example reasoning trace in Figure~\ref{fig:explicit_reasoning_trace}.

\end{document}